\documentclass[twoside]{article}

\usepackage[accepted]{aistats2023}
\usepackage[round]{natbib}

\usepackage{xcolor}
\usepackage{graphicx}
\usepackage{caption}
\usepackage{enumerate}
\usepackage{amsmath}
\usepackage{amssymb}
\usepackage{hyperref}
\usepackage{amsthm}
\usepackage{algorithm}
\usepackage{algpseudocode}
\usepackage{subcaption}
\usepackage{booktabs}
\newtheorem{corollary}{Corollary}
\newtheorem{remark}{Remark}
\newtheorem{lemma}{Lemma}
\theoremstyle{definition}
\newtheorem{definition}{Definition}
\newtheorem{fact}{Fact}

\newcommand \E {\mathop{\mbox{\ensuremath{\mathbb{E}}}}\nolimits}

\renewcommand \Pr {\mathop{\mbox{\ensuremath{\mathbb{P}}}}\nolimits}

\newcommand{\tuple}[1]{\left\langle #1\right\rangle }

\newcommand{\cset}[2]{\left\{\, #1 ~\middle|~ #2 \,\right\} }

\newcommand\Simplex {{{\Delta}}}
\newcommand\Reals {{\mathbb{R}}}

\newcommand \CA {{\mathcal{A}}}

\newcommand \CH {{\mathcal{H}}}

\newcommand \CM {{\mathcal{M}}}

\newcommand \CS {{\mathcal{S}}}

\newcommand \bw {{\mathbf{w}}}

\newcommand \p[1] {\frac{\partial}{\partial #1}}

\newcommand \defn {\mathrel{\triangleq}}
\newcommand \argmax{\mathop{\rm arg\,max}}

\newcommand \supp{\mathop{\rm supp}}

\newcommand \trans[1]{#1^{\top}}

\newcommand \grad {\nabla}

\DeclareMathAlphabet{\mathpzc}{OT1}{pzc}{m}{it}

\newcommand \Beta {\mathop{\mathpzc{Beta}}\nolimits}

\newcommand \bel {\beta}
\newcommand \Bels {\mathcal{B}}
\newcommand \pure {d}
\newcommand \str {\sigma}
\newcommand \pol {\pi}
\newcommand \Pols {\Pi}
\newcommand \PolsHD {\Pols^{\mathrm{D}}}
\newcommand \PolsMD {\Pols_1^{\mathrm{D}}}
\newcommand \PolsHS {\Pols^{\mathrm{S}}}
\newcommand \PolsMS {\Pols_1^{\mathrm{S}}}

\newcommand \PolsEps {\Pols^\epsilon}

\newcommand \mdp {\mu}
\newcommand \MDPs {\CM}

\newcommand \sutil {{u}}
\newcommand \util {{U}}
\newcommand \regret {{R}}
\newcommand \eregret {{L}}

\newcommand \Ebp {\E_\bel^\pol}

\newcommand \rew {\rho}

\newcommand\dd{\,\mathrm{d}}

\begin{document}


\runningauthor{Thomas Kleine Buening, Christos Dimitrakakis, Hannes Eriksson, Divya Grover, Emilio Jorge}

\twocolumn[

\aistatstitle{Minimax-Bayes Reinforcement Learning}

\aistatsauthor{Thomas Kleine Buening$^*$ \\ University of Oslo \And Christos Dimitrakakis$^*$  \\ University of Neuchatel \And Hannes Eriksson$^*$  \\ Zenseact \AND Divya Grover$^*$  \\ Chalmers University of Technology \And Emilio Jorge$^*$  \\ Chalmers University of Technology 
\vspace{0.1cm}}

\aistatsaddress{ } ]

\begin{abstract}
  While the Bayesian decision-theoretic framework offers an elegant
  solution to the problem of decision making under uncertainty, one
  question is how to appropriately select the prior distribution. One
  idea is to employ a worst-case prior. However, this is not as easy to
  specify in sequential decision making as in simple statistical
  estimation problems.  This paper studies (sometimes approximate)
  minimax-Bayes solutions for various reinforcement learning problems
  to gain insights into the properties of the corresponding priors and
  policies. We find that while the worst-case prior depends on the
  setting, the corresponding minimax policies are more robust than
  those that assume a standard (i.e.\ uniform) prior.
\end{abstract}

\section{Introduction}
\label{sec:introduction}

Reinforcement learning is the problem of an agent learning how to act
in an unknown environment through interaction and reinforcement. In
the standard setting, the learning agent acts in an unknown Markov
Decision Process $\mdp$, within some class of MDPs $\MDPs$. The agent
observes the state $s_t \in \CS$ of the MDP and selects an action $a_t \in \CA$
using a policy $\pol$. It then observes a reward $r_t \in \Reals$ and
the next state $s_{t+1}$. The agent's goal is to maximise utility,
defined as the sum of rewards to some horizon $T$,
$\sutil = \sum_{t=1}^T r_t$, in expectation, i.e.\
$\E^\pol_ \mdp(\sutil)$, where $\E^\pol_\mdp$ is the expectation under
the MDP and policy. Since the true $\mdp$ is unknown, this
optimisation problem is ill-posed. In the Bayesian setting, this
conundrum is solved by selecting some \emph{subjective} prior
distribution $\bel$ over MDPs and maximising
$\E^\pol_\bel(\sutil) = \int_\MDPs \E^\pol_\mdp(\sutil) \dd
\bel(\mdp)$.  Then it remains to compute the optimal adaptive
(i.e. history-dependent) policy, something that can be only done
approximately in general, due to the fact that the number of adaptive
policies increases exponentially with the problem horizon.
 
The above discussion assumes that the agent has \emph{somehow} chosen
a prior. However, it is not clear how such a prior can be selected
from first principles, if we have no domain knowledge, but
still want to be robust.  The minimax-Bayes
idea~\citep{berger:statistical-decision-theory} is to assume that
nature selects the \emph{worst} possible prior $\bel^*$ for the agent,
but \emph{without} knowledge of the agent's policy. This can be
formalised by having nature play the minimising player in a
simultaneous-move zero-sum game defined by the expected utility
$\E^{\pol}_\bel(\sutil)$, where the agent (who maximises) chooses
$\pol$, and nature (who minimises) chooses $\bel$.  
In simple Bayesian
decision problems (e.g.\ linear regression) the minimax-Bayes problem
is well-studied and $\bel^*$ sometimes corresponds to a maximum
entropy prior. However, in an interactive setting, results are limited
to one-shot experiment
design~\citep{grunwald-dawid:game-robust-bayesian:aos:2004}, which
shows that maximum entropy priors are not the worst-case priors
generally.

In reinforcement learning, which can be seen as a sequential
generalisation of one-shot experiment design, this problem has not
received much attention in the past. Sometimes, the concept of maximum
entropy has been used in reinforcement learning as a penalty term on
the policy~\citep[e.g.][]{todorov2006linearly, haarnoja2018soft,
  eysenbach2021maximum} as well as in the context of inverse
reinforcement learning~\citep{ziebart2010modeling}, but an explicit
connection to the minimax-Bayes literature has not been made.  In
preliminary work, \cite{androulakis2014generalised} analysed
variants of the weighted majority algorithm for finding minimax priors
in a restricted version of this setting.

\textbf{Contributions.} In this paper, we study the basic theoretical
and algorithmic properties of minimax-Bayes reinforcement
learning. This includes (a) characterising the existence of solutions
under different assumptions on the policy and MDP space (b) defining
algorithms, together with convergence guarantees when possible, and
(c) performing numerical experiments to illustrate the behaviour of
(approximate) minimax-Bayes algorithms and contrast them with Bayesian RL algorithms
that assume a standard maximum-entropy (e.g.\ uniform) prior.

The paper is organised as follows. In Section~\ref{sec:setting}, we
formally introduce the setting.  In Section~\ref{sec:regret}, we
introduce regret definitions and prove some basic properties of the
regret as well as relations between Bayesian regret and Bayes-optimal
regret. Section~\ref{sec:minimax-theorems} discusses the existence of
a value for the game between a Bayesian agent and Nature, which selects
the prior.  Section~\ref{sec:algorithms} develops algorithms for
finding approximately minimax policies in certain policy classes. In
particular, we consider (a) finite-horizon Bayes-optimal policies (b)
posterior sampling policies, and (c) parametrised adaptive
policies. Our results indicate that, not only is an approximate
minimax solution achievable in many settings but that they are much
more robust than Bayes-adaptive policies under common priors. Finally, Section \ref{sec:conclusion} contains the related work and conclusions.

\section{Setting}
\label{sec:setting}
A Markov Decision Process (MDP) is a tuple
$\mu = \tuple{\CS, \CA, P, \rew, T}$, where $\CS$ is a set of states,
$\CA$ is a set of actions, $P: \CS \times \CA \to \Simplex(\CS)$ is a
transition function,
%, such that $s_{t+1} \mid s_t = s, a_t = a \sim P^{s,a}$,
$\rew: \CS \times \CA \to [0,1]$ is a reward function, and $T$ is a
(potentially random) horizon.  
Let $\MDPs$ denote the space of MDPs.

For simplicity, in our theoretical development, we focus on the setting
where the agent is acting in a finite state space $\CS$ with a finite
set of actions $\CA$, the reward function $\rho$ is known, and the
horizon $T$ is fixed and finite, although many of our results could
be more generally applicable.  In each round $t$, the agent observes
state $s_t\in \CS$, chooses an action $a_t\in \CS$ and receives a
reward $r_t = \rho(s_t, a_t)$. We write $s^t = (s_1, \dots, s_t)$ and
$a^t = (a_1, \dots, a_t)$ for the sequence of states and actions up to
round $t$. Given the reward function, the history
$h_t = (s^t, a^{t-1})$ describes the information available to
the agent before choosing an action in round~$t$.  The agent's utility
$\sutil$ is an additive function of individual rewards
$\sutil \defn \sum_{t=1}^T r_t$. The agent is acting in an MDP through
a policy $\pi \in \Pi$, where we let $\Pi$ denote a generic policy
space. For a fixed MDP $\mdp \in \MDPs$ and policy $\pi \in \Pi$, the
expected utility is given by
$\util(\pi, \mdp) \defn \E^\pi_\mdp [\sutil]$ with maximal utility
denoted by $\util^*(\mdp) \defn \max_{\pi \in \Pi} \util(\pi, \mdp)$.

When the MDP is unknown, as in the reinforcement learning problem, the policy is adaptive and the agent's actions can depend on what it
has been observed in the past, as we explain below.

%We first recall that a Markov decision process $\mdp \in \MDPs$ on a state-action space $\CS \times \CA$ is a tuple
%$\tuple{\CS, \CA, P, \rew, T}$ where $\CS$ is a set of states, $\CA$
%is a set of actions, $P$ is a transition kernel, such that
%$s_{t+1} \mid s_t = s, a_t = a \sim P^{s,a}$, where
%$r_t = \rew(s_t,a_t)$ and $T$ is a (potentially random) horizon.  Here
%we focus in the setting where the agent is acting in a finite state
%space $\CS$ with a finite set of actions $\CA$. The horizon itself is
%also finite. For simplicity, in the remainder we will assume that
%$r_t \in [0,1]$.  The agent's utility $\util$ is an additive function
%of individual rewards $r_t \in \Reals$, with
%$\util \defn \sum_{t=1}^T r_t$.  The agent is acting in an MDP through
%a (possibly adaptive) policy.

\subsection{Policies.}
Let $\CH$ be the set of all histories. %$(s^t, a^{t-1})$. 
A (stochastic) policy $\pol$ is a set of probability measures
$\cset{\pol(\cdot \mid h)}{h \in \CH}$ on the set of actions $\CA$. We
denote the set of all behavioural\footnote{That is, history-dependent
	and stochastic policies.} policies by $\PolsHS$.  A policy is
\emph{deterministic} if, for each history $h_t=(s^t, a^{t-1})$, there
exists an action $a \in \CA$ such that
$\pol(a_t = a \mid h_t) = 1$. We denote the set of
deterministic policies by $\PolsHD$.  A policy is \emph{memoryless}
(or reactive) if, for all histories $h_t$ with $s_t = s$, we have
$\pol(a_t = a\mid h_t) = \pol(a_t = a \mid s_t = s)$. We denote
the set of memoryless (stochastic) policies by $\PolsMS$.  The set of
memoryless deterministic policies is denoted by $\PolsMD$.  Obviously,
$\PolsMD \subset \PolsHD \subset \PolsHS$ and
$\PolsMD \subset \PolsMS \subset \PolsHS$.  Finally, for any MDP
$\mdp$ there exists a deterministic, memoryless policy that is
optimal, i.e.
$\util^*(\mdp) = \max_{\pol \in \Pols} \util(\pi, \mdp) = \max_{\pol
	\in \PolsMD} \util(\pi, \mdp)$ \citep[see e.g.][]{puterman2014markov}.

\paragraph{Strategies.} 
Typically, minimax results rely on the notion of mixed strategies. 
Here, we let $\str \in \Simplex(\Pols)$ denote a probability measure over a set of base policies $\Pols$. 
\begin{fact}
	For any strategy $\str \in \Simplex(\PolsHD)$ there exists an equivalent stochastic policy $\pol \in \PolsHS$ such that $\str(a_t | h_t) = \pol(a_t \mid h_t)$ for all histories $h_t$ with positive probability.
	\label{fact:mixed-stochastic-equivalence}
\end{fact}

\subsection{Utility and Beliefs}
\label{sec:utility}
In the following, we overload the $U(\pol, \bel)$ to also mean the expected utility of $\pol$ with respect to a distribution $\bel$ over MDPs:
\begin{equation}
  \label{eq:utility}
  \util(\pol, \bel) \defn \Ebp [\sutil]
  = \int_\MDPs \util(\pi, \mdp) \dd{\bel}(\mdp),
\end{equation}
under appropriate measurability assumptions.
%For a fixed MDP $\mdp \in \MDPs$ and policy $\pol \in \Pols$, the
%expected utility is $\E^\pol_\mdp \util$, while the conditional
%expected utility is called the \emph{value function}:
%$V^\pol_\mdp(s) \defn \E_\mdp^\pol \left(\sum_{t=1}^T r_t ~\middle|~
%  s_t = s \right)$.  We additionally define
%$V^\pol_\bel \defn \int_\MDPs V^\pol_\mdp \dd{\bel}(\mdp)$ for the
%value function under a distribution $\bel$ on the MDPs.  Finally, for
%a given probability measure $\init$ on $\CS$, which can be taken to
%represent a starting state distribution, we define the utility of a
%particular policy $\pol$ to be:
%\begin{equation}
%  \label{eq:utility}
%  \util(\bel, \pol) \defn \Ebp \init^\top V^\pol_\mdp
%  = \int_\MDPs \init^\top V^\pol_\mdp \dd{\bel}(\mdp).
%\end{equation}

There are two possible ways to interpret the distribution $\bel$, depending
on how it is chosen. If $\bel$ is chosen by the agent selecting
$\pol$, it corresponds to the subjective belief of the decision
maker about which is the most likely MDP \emph{a priori}. Then,
$\util(\pol, \bel)$ corresponds to the expected utility of a
particular policy under this belief.  Let
$$\util^*(\bel) \defn \max_{\pol \in \Pols} \util(\pol, \bel)$$ denote
the Bayes-optimal utility for a belief.  We recall the fact that this
is a convex
function~\citep[c.f.][]{Degroot:OptimalStatisticalDecisions}. By
definition, the following bounds hold:
\[
  \util(\pol, \bel) \leq \util^*(\bel) \leq \int_\MDPs \util^*(\mdp)
  \dd{\bel}(\mdp)
  ,
  \qquad \forall \pol \in \Pols,
\]
so that $U^*(\bel)$ is
convex with respect to $\bel$. In the above, the left-hand side is the
utility of an arbitrary policy, while the right side can be seen as
the expected utility we would obtain if the true MDP was revealed to
us.

The second view of $\bel$ is to assume that the MDP is \emph{actually}
drawn randomly from the distribution $\bel$. If this is known, then
the subjective value of a policy is equal to its true expected
value. However, it is more interesting to consider the case where
nature arbitrarily selects $\bel$ from a set of possible
priors $\Bels$. Then we wish to find a policy $\pol^*$ achieving:
\begin{equation}
	\label{eq:maximin-policy}
	\max_{\pol \in \Pols} \min_{\bel \in \Bels} \util(\pol, \bel).
\end{equation}
A minimax solution exists if the game \emph{has a value}, i.e.\
$\max_{\pol \in \Pols} \min_{\bel \in \Bels} \util(\pol, \bel) =
\min_{\bel \in \Bels} \max_{\pol \in \Pols} \util(\pol, \bel)$.  
Then there exists a maximin policy $\pol^*$ which is
optimal in response to some minimax belief $\bel^*$, and vice versa. A
sufficient condition for this to occur is for $\util^*(\bel)$ to be
convex and differentiable
everywhere~\citep[c.f.][]{grunwald-dawid:game-robust-bayesian:aos:2004}.
In particular, a maximin \emph{strategy} (i.e.\ a distribution over policies) can always be found when $\Pols$
is finite. On the other hand, for any fixed prior $\bel$, there is always an optimal deterministic policy. Note that this is only a \emph{best-response} policy and not a solution to
the maximin problem~\eqref{eq:maximin-policy}.
\begin{fact}
	For any distribution $\bel$ over MDPs, there exists a deterministic, history-dependent policy that is optimal, i.e.
	$\util^*(\bel) = \max_{\pol \in \Pols} \util(\pol, \bel) = \max_{\pol \in \PolsHD} \util(\pol, \bel)$. 
\end{fact}
      
Unfortunately, looking at the problem from the point of view of
utility maximisation is somewhat problematic. This is because an
unrestricted set of priors for nature may lead to absurd solutions:
nature could pick a prior so that all rewards are zero, thus trivially
achieving minimal utility. For that reason, we actually focus on the
problem of minimax \emph{regret}, i.e.\ the gap between the agent's
policy and that of an oracle. We give the appropriate definitions in
the next section.

\section{Properties of the regret}
\label{sec:regret}
We generally write $\regret(\pol, \mathcal{I})$ to mean the regret of
some algorithmic policy $\pol$ relative to an oracle with information~$\mathcal{I}$.
%For notational simplicity, in the following we consider
%a finite set of base MDPs $\MDPs$.

Let us start with the regret of a policy relative
to an oracle that knows the underlying MDP:
\begin{definition}[Regret] The regret of a policy $\pi$ for an MDP $\mdp$ is $\regret(\pol, \mdp) \defn \util^*(\mdp) - \util(\pol, \mdp)$.
\end{definition}
Since this regret notion may be too strong, it is also interesting to define the regret of a policy with respect to the oracle that knows $\bel$. This allows us to take into account oracles which have less knowledge than the actual MDP.
\begin{definition}[Bayes-optimal Regret] This is the regret of a policy $\pol$ with respect to the Bayes-optimal policy\footnote{Generally this policy will belong to the set of history-dependent policies, but in some cases, it makes sense to restrict them to e.g. a subset of parametrised policies.} for $\bel$:
  $\regret(\pol, \bel) \defn \util^*(\bel) - \util(\pi, \bel)
  =
  \int_\MDPs \dd \bel(\mdp) [\util(\pol^*(\bel), \mdp) - \util(\pol, \mdp)]$,
  where $\pol^*(\bel) = \arg\max_{\pol} \util(\pol, \bel)$.
\end{definition}
This notion of regret tells us how much we lose relative to a computationally unbounded oracle that knows the prior. We can use it
to measure the loss both due to a misspecified prior, by fixing
$\pol^*(\bel_0)$ to some prior $\bel_0$ and examining
$\regret(\pol^*(\bel_0), \bel)$ as the actual prior $\bel$ varies, and
due to computational approximations, by measuring
$\regret(\pol^*_\epsilon(\bel), \bel)$ for policies calculated with
some approximate algorithm.
      
Finally, we may wish to subjectively calculate our expected regret
under an oracle that knows the underlying MDP. Since the agent does
not know the underlying MDP, it necessarily measures regret under a
Bayesian prior.

\begin{definition}[Bayesian regret]
	The Bayesian regret of a policy $\pol$ under a prior $\bel$ is
	$\eregret(\pol, \bel) \defn \E_{\mdp \sim \bel} [\regret(\pol, \mdp)] =  \sum_\mdp \bel(\mdp) \regret(\pol, \mdp) =  \sum_\mdp \bel(\mdp) [\util^*(\mdp) - \util(\pol, \mdp)]$.
\end{definition}
These definitions of regret are closely related, as we shall show
in the remainder.  It will be illuminating to look at the difference
between the regret the agent subjectively expects to suffer with
respect to some prior distribution $\bel$, relative to the regret of
the same policy compared to the Bayes-optimal policy for the same
prior. 
% An annotated illustration of these notions of regret is given in Figure \ref{fig:regret}.

 \begin{figure}[h]
	\centering
	\includegraphics[width=0.85\linewidth]{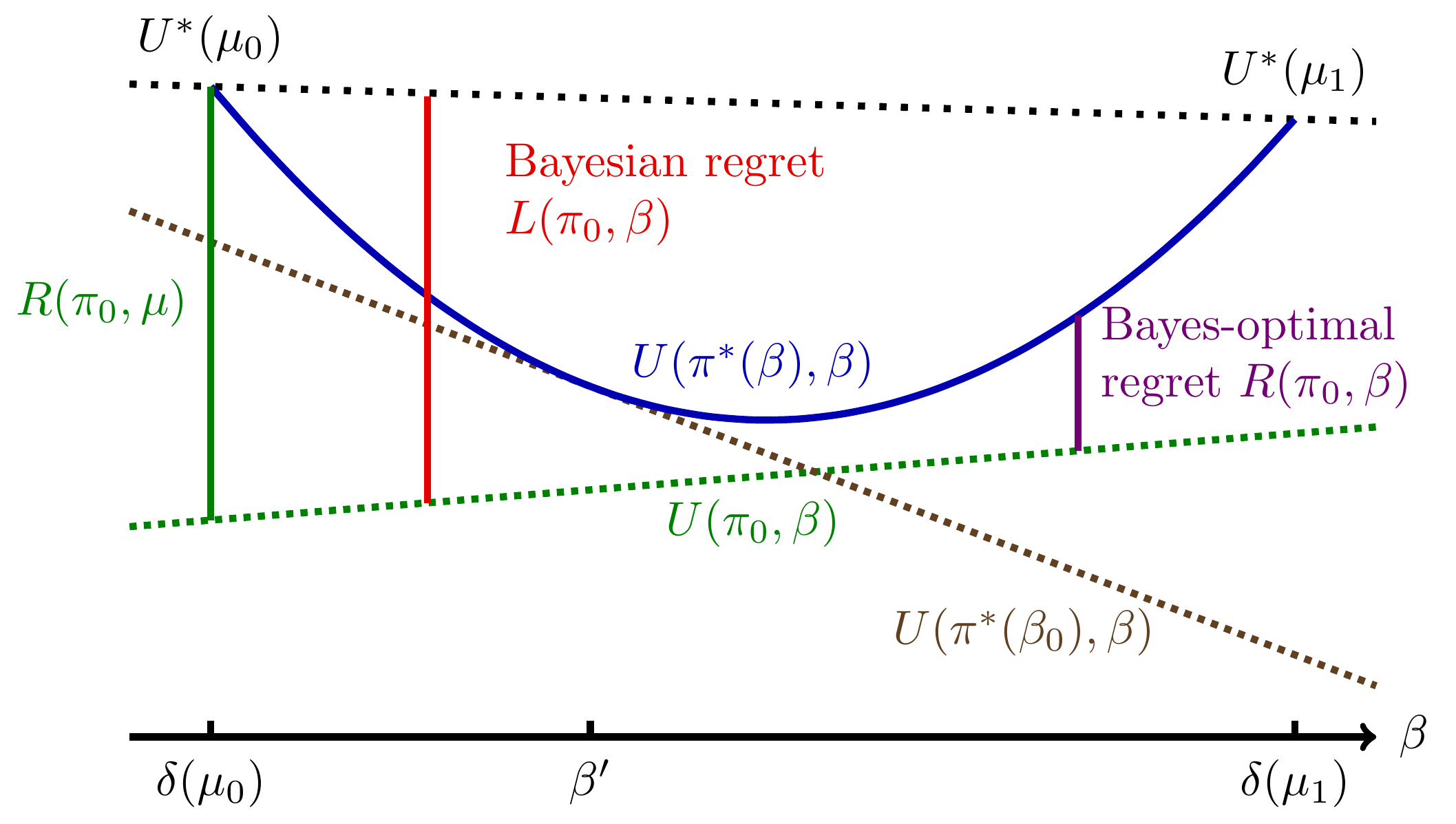}
	\caption{Illustration of the notions of regret for different
      policies with a belief $\bel$ over two MDPs $\mu_1$ and $\mu_2$, where $\delta(\mdp)$
      denotes the Dirac belief on $\mdp$.  Any \emph{fixed} policy
      $\pi_0$ will have a utility that is a linear function of the
      belief (green dotted line). The blue curve shows the utility of the Bayes-optimal
      policy $\pi^*(\beta)= \argmax_\pi \util(\pi, \beta)$. This
      policy is prior-aware, and hence not fixed, but depends on the
      prior $\beta$. Note that by definition, $\util(\pi^*(\beta), \beta)$ is
      convex. However, if we fix a Bayes-optimal policy for a specific
      prior~$\beta_0$, we obtain a tangent
      $\util(\pol^*(\beta_0), \beta)$ to the Bayes-optimal curve at
      $\beta_0$. The Bayesian regret (of $\pi_0$) (red line) is the expected regret of a
      policy compared against an oracle that knows the MDP (black dotted line). The
      Bayes-optimal regret (of $\pi_0$) is the difference in performance to the
      Bayes-optimal policy (purple line).}
		\label{fig:regret}
\end{figure}

\begin{remark}
	The Bayesian regret of a policy $\pol$ is greater than the Bayes-optimal regret, i.e.\
	$\regret(\pol, \bel) \leq \eregret(\pol, \bel)$.
\end{remark}
\begin{proof}
  Note that
  $\regret(\pol, \bel) = \int_\MDPs \dd \bel(\mdp) [\util(\pol^*(\bel),
  \mdp) - \util(\pol, \mdp)] \leq \int_\MDPs \dd \bel(\mdp) [\util^*(\mdp)
  - \util(\pol, \mdp)] = \eregret(\pol, \bel)$, since
  $\util(\pi^*(\beta), \mdp) \leq \util^*(\mdp)$ by definition of
  $\util^*(\mdp)$.  
\end{proof}
The above also follows from the fact that for any policy $\pol$ and
prior $\bel$, the Bayesian regret of $\pi$ equals the Bayesian regret
of the Bayes-optimal policy\footnote{This is equal to the difference
  between the Bayes-optimal value and the upper bound.} plus the
Bayes-optimal regret of $\pol$, that is,
$\eregret(\pol, \bel) = \eregret(\pol^*(\bel), \bel) + \regret(\pol,
\bel)$. Geometrically, this follows from the fact that the utility of
any fixed policy is lower bounding the convex Bayes-optimal utility
curve, as can be seen in Figure~\ref{fig:regret}.
The following fact also follows from a simple geometrical argument:
\begin{remark}
	$\regret(\pol, \bel)$ is convex in $\bel$.
	\label{rem:convex-regret}
\end{remark}
\begin{proof}
	By definition of the Bayesian-optimal regret, we have $\regret(\pol, \bel) = \util^*(\bel) -  \E_{\mdp \sim \bel} [\util(\pol, \mdp)]$. As $\util^*(\bel)$ is convex in $\bel$ and $\E_{\mdp \sim \bel} [\util(\pol, \mdp)]$ is linear in $\bel$, their difference is also convex.
\end{proof}
Of course, the game where nature sees the agent's policy $\pol$ first
before selecting a prior is strictly determined and nature can simply
select a single MDP (Dirac distribution) as its best response to $\pol$. In this particular case, this
follows directly from the convexity of the Bayes-optimal regret.

%Intuitively, the worst-case regret of any policy can be taken over MDPs rather than beliefs, as the Bayes-optimal regret is a convex function. This implies that, for any policy, the maxima of the function lie on beliefs which are degenerate.
% Formally, letting $\supp(\bel)$ denote the support of a distribution $\bel$ and $\supp(\Bels) = \bigcup_{\bel \in \Bels} \supp(\bel)$ denote the support of a set of distributions, it is possible to show (see Appendix), following the steps of the proof by~\cite{tor:pc} for the bandit case, that maximum regret is attained in degenerate beliefs. 
Following the steps of the proof by~\cite{tor:pc} for the bandit case, we can show that the maximum regret is attained in Dirac beliefs. Here, we let $\Bels$ denote the set of beliefs and we work under the assumption that the degenerate beliefs are contained in the belief space.  
%\todo[inline]{This should not need convexity?}
\begin{lemma}[\textnormal{\cite{tor:pc}}]
	% If $\supp(\Bels) = \MDPs$ and for each MDP $\mdp$ there exists an associated Dirac belief $\bel_\mdp$, then
	If for each MDP $\mdp\in \MDPs$ there exists an associated Dirac belief $\bel_\mdp \in \Bels$, then for any policy $\pi$ we have 
	$\max_{\mdp \in \MDPs} \regret(\pol, \mdp) = \max_{\bel \in \Bels} \regret(\pol, \bel)$.
	\label{lem:tor-pc}
\end{lemma}
This immediately implies that the minimax regret is the same over both beliefs and MDPs: 
\begin{equation}
	\min_{\pol \in \Pi} \max_{\mdp \in \MDPs} \regret(\pol, \mdp) =   \min_{\pol \in \Pi} \max_{\bel\in \Bels} \regret(\pol, \bel)
	\label{eq:minimax-regret}
\end{equation}
%We can also prove a similar result, with a different proof method, for the Bayesian regret. 
We find a similar result for the Bayesian regret. 
\begin{lemma}\label{lem:max_dirac_bayesian_regret}
	If for each MDP $\mdp\in \MDPs$ there exists an associated Dirac belief $\bel_\mdp \in \Bels$, then for any $\pi$:
	\begin{equation}
		\max_{\mdp \in \MDPs} \regret(\pol, \mdp) = \max_{\bel \in \Bels} \eregret(\pol, \bel).
		\label{eq:eregret-equivalence}
	\end{equation}
\end{lemma}
\begin{proof}
	For any $\bel  \in \Bels$, we have
	\begin{align*}
		\max_{\mdp \in \MDPs} \regret(\pol, \mdp)  & \geq \max_{\mdp \in \supp(\bel)} \regret(\pol, \mdp) \\
		&  = \max_{\mdp \in \supp(\bel)} \util(\pol^*(\mdp), \mdp) - \util(\pol, \mdp) \\ 
		& \geq \int_{{\supp(\bel)}} \dd \bel(\mdp) [\util(\pol^*(\mdp), \mdp) - \util(\pol, \mdp)] \\
		&= \eregret(\pol, \bel).
	\end{align*}
	% $\max_{\mdp \in \MDPs} \regret(\pol, \mdp) \geq \max_{\mdp \in \supp(\bel)} \regret(\pol, \mdp) = \max_{\mdp \in \supp(\bel)} \util(\pol^*(\mdp), \mdp) - \util(\pol, \mdp) \geq \sum_{\mdp \in \supp(\bel)} \bel(\mdp) [\util(\pol^*(\mdp), \mdp) - \util(\pol, \mdp)] = \eregret(\pol, \bel)$.

	Consequently $\max_\mdp \regret(\pol, \mdp) \geq \max_\bel \eregret(\pol, \bel)$. Using $\delta(\MDPs)$ to denote the set of Dirac beliefs over $\MDPs$, we have:
	$\max_\bel \eregret(\pol, \bel)
	\geq
	\max_{\bel \in \delta(\MDPs)} \eregret(\pol, \bel) = 
	\max_{\mdp \in \MDPs} \regret(\pol, \mdp)$,
	due to the fact that $\regret(\pol, \mdp) = \eregret(\pol, \bel_\mdp)$ for the singular belief $\bel_\mdp$ on MDP $\mdp$.
	As a result, it must hold that $\max_{\mdp \in \MDPs} \regret(\pol, \mdp) \geq \max_{\bel \in \Bels} \eregret(\pol, \bel) \geq \max_{\mdp \in \MDPs} \regret(\pol, \mdp)$. 
\end{proof}

\citet{lattimore2019information} show that for the problem of prediction with partial information, the minimax regret equals the minimax Bayesian regret. We show that this also holds in a general setting, as an immediate consequence of Lemma~\ref{lem:max_dirac_bayesian_regret}.
%\todo{only directly parameterized?}
\begin{corollary}
	If for each MDP $\mdp\in \MDPs$ there exists an associated Dirac belief $\bel_\mdp \in \Bels$, then for any $\pi$:
	\begin{equation}
		\min_{\pol \in \Pi}  \max_{\mdp\in\MDPs} \regret(\pol, \mdp) = \min_{\pol \in \Pi} \max_{\bel\in \Bels} \eregret(\pol, \bel)
		\label{eq:minimax-bayes-regret}
	\end{equation}
\end{corollary}
%\begin{proof}
%  For any $\pol$, equation~\eqref{eq:eregret-equivalence} holds, so $\min_\pol \max_\bel \regret(\pol, \bel) = \min_\pol \max_\mdp \regret(\pol, \mdp)$. More precisely, if $\pol^*$ is a policy minimising $\max_\bel \regret(\pol, \bel)$ then $\max_\mdp \regret(\pol^*, \mdp) = \max_\bel \regret(\pol^*, \bel) \leq \max_\bel \regret(\pol, \bel) = \max_\mdp \regret(\pol, \mdp)$ for all $\pol$, so it is also minimising $\max_\mdp \regret(\pol, \mdp)$.
%\end{proof}
Equations \eqref{eq:minimax-regret} and  \eqref{eq:minimax-bayes-regret} can be made intuitive through a simple geometric argument.  Due to the linearity of the expected regret with respect to the belief for any fixed policy, the best response for nature always includes singular beliefs.

\section{Minimax theorems}
\label{sec:minimax-theorems}
The above results merely make precise the intuition that when playing second, nature does not need to randomise: it can simply pick the worst-case MDP for the policy we have chosen. However, we typically want to model a worst-case setting by assuming nature picks its distribution without knowing which policy the decision maker will pick. For that reason, it is important to investigate whether the normal form game against nature, where nature and the agent play without seeing each other's move, has a value. We would expect this to be the case if the regret was a bilinear function of the policy and prior. Consequently, the answer is positive with respect to both the Bayesian regret and the utility in the finite setting. However, this is not the case for the Bayes-optimal regret.
\begin{corollary}\label{cor:minimax_bayesian_regret}
	For a finite set of MDPs in a finite state-action space, with a known reward function and a finite horizon, the utility and Bayesian regret satisfy:
	\begin{align}
		\min_{\bel \in \Bels} \max_{\pol \in \Pols} \util(\pol, \bel)
		&=
		\max_{\pol \in \Pols} \min _{\bel \in \Bels} \util(\pol, \bel), \\
		\qquad
		\max_{\bel \in \Bels} \min_{\pol \in \Pols}  \eregret(\pol, \bel)
		&=
		\min_{\pol \in \Pols} \max _{\bel \in \Bels} \eregret(\pol, \bel)
	\end{align}
\end{corollary}
\begin{proof}
	First note that, due to Fact~\ref{fact:mixed-stochastic-equivalence}, the stochastic policy $\pol$ can always be written as a distribution $\str$ over deterministic behavioural policies $\pure \in \PolsHD$ so that 
	$\util(\pol, \bel)
	=
	\sum_{\mdp} \sum_{\pure} \bel(\mdp) \util(\pure, \mdp) \str(\pure)$.
	The result follows from the standard minimax theorem. Similarly for regret, we use
	$\eregret(\pol, \bel)
	=
	\sum_{\mdp} \sum_{\pure} \bel(\mdp) \regret(\pure, \mdp) \str(\pure)$.
\end{proof}
The same does not hold for the Bayes-optimal regret, since for arbitrary policy spaces the agent's Bayes-optimal policy has zero Bayes-optimal regret, as it is aware of the prior distribution. However, the minimax value is generally greater than zero.
%Since $\regret(\pol, \bel)$ is non-linear in $\bel$, we do not obtain the standard bilinear form, and the game may not have a value.
%\todo[inline]{Again, R not convex?}
% I don't know what the point of that line was, but for fixed pi, 
\begin{lemma}
	The game $\regret(\pol, \bel)$ does not have a value when $\MDPs$ contains at least two MDPs $\mdp, \mdp'$ whose optimal policy sets have an empty intersection.
\end{lemma}
\begin{proof}
	For $\pol \in \PolsHD$, we have $\max_\bel \min_\pol \regret(\pol, \bel) = 0$, so that
	$\min_\pol \max_\bel \regret(\pol, \bel) \geq \max_\bel \min_\pol \regret(\pol, \bel)~=~0$.
	From~\eqref{eq:minimax-regret}, it then follows that
	$\min_\pol \max_\mdp \regret(\pol, \mdp) = \min_\pol \max_\bel \regret(\pol, \bel) \geq \max_\bel \min_\pol \regret(\pol, \bel) = 0$.
	It remains to show that
	$\min_\pol \max_\mdp \regret(\pol, \mdp) > 0$. Assume the
	contrary. Then there is some policy $\pol^*$ for which
	$\max_\mdp \regret(\pol^*, \mdp) =0 $. However, there exists at
	least one $\mdp'$ whose optimal policy does not coincide with
	$\pol^*$, hence $\regret(\pol^*, \mdp') > 0$.
\end{proof}
Finally, it is interesting to consider the Bayesian regret of the Bayes-optimal policy. For the worst-case Bayesian regret of the Bayes-optimal policy, we find that it is equal to the minimax Bayesian regret. 
\begin{lemma} 
  For finite $\MDPs$, the worst-case Bayesian regret of the
  Bayes-optimal policy equals the minimax Bayesian regret, i.e.\
	\[
	\max_{\bel \in \Bels} \eregret(\pol^*(\bel), \bel) = \max_{\bel \in \Bels} \min_{\pol \in \Pols} \eregret(\pol, \bel) = \min_{\pol \in \Pols} \max_{\bel \in \Bels} \eregret(\pol, \bel).  
	\]
\end{lemma}
\begin{proof}
  % [Note that always: $\max_\bel \eregret(\pol^*(\bel), \bel) \geq \max_\bel \min_\pol \eregret(\pol, \bel)$.]
	By definition of the Bayes-optimal policy, we have $\util(\pol^*(\bel), \bel) = \max_\pol \util(\pol, \bel)$. Thus, 
	\begin{align*}
		\max_\bel \eregret(\pol^*(\bel), \bel)
		& = \max_\bel \sum_{\mdp} \bel(\mdp) [\util^*(\mdp) - \util(\pol^*(\bel), \mdp)] \\ 
		& = \max_\bel \min_\pol \sum_\mdp \bel(\mdp) [ \util^*(\mdp) - \util(\pol, \mdp)]\\
		& = \max_\bel \min_\pol \eregret(\pol, \bel). 
	\end{align*}
	While the above holds for arbitrary $\MDPs$, for the second equality we need to use Corollary~\ref{cor:minimax_bayesian_regret}, which states that the game has a value when $\MDPs$ is finite, so that $\max_\bel \min_\pol \eregret(\pol, \bel) = \min_\pol \max_\bel \eregret(\pol, \bel)$. 
\end{proof}
It is important to emphasise that this does not imply that $\pol^*(\bel^*)$ is a minimax policy, but merely that its value at the worst-case belief $\bel^*$ is equal to the value of the game. As we shall see in Section~\ref{sec:finitemdp}, in settings with a finite number of policies, $\bel^*$ is located at a vertex with at least two best response policies $\pol^*$, where the minimax policy must be a mixture between those.

\paragraph{Open questions.} This concludes our preliminary discussion of minimax values for Bayesian games on MDPs. While it is clear that standard minimax theorems apply in the discrete case when we consider stochastic policies, it is an open question whether those can be extended to a more general setting. In particular, do the utility and Bayesian regret game have a value with an uncountable family of priors such as the Dirichlet-product prior? It is also an open question whether a value for the game exists when we are restricted to deterministic policies in some cases. We conjecture that this is generally not the case. For example in discrete, finite horizon problems, the set of policies pure deterministic policies is finite, and so it is unlikely that one of them is maximin. We explore these questions experimentally, after we first develop some algorithms in the following section.

\section{Algorithms}
\label{sec:algorithms}
In this section, we attempt to answer some of the above questions empirically. In particular, does there exist an equilibrium for bandit problems, where the Bayes-optimal policy can be efficiently approximated through Gittins indices? What about settings where we must restrict the policy space to parametrised or tree policies? Does solving the minimax problem approximately lead to robust policies? Are the worst-case priors we obtain through optimisation actually preferable in some way to standard priors such as the uniform one? For example, do they lead to more robust policies? 

For the infinite horizon case, we cannot consider the Bayes-optimal
regret, as it requires us to compute the Bayes-optimal
policy. However, we can always target the Bayesian regret, which is an
upper bound on the Bayes-optimal regret. (And since the former is
usually the same as the minimax regret, it gives us a minimax
policy). 

Section~\ref{sec:gradient-methods} describes a stochastic
gradient descent-ascent algorithm for finding an approximate minimax
regret pair. For the finite horizon case, we can obtain the
Bayes-optimal response to any prior distribution. More specifically,
when the set of possible MDPs is finite, and we have an optimal policy
oracle, we can employ a cutting plane algorithm, described in
Section~\ref{sec:cutting-planes}. This allows us to obtain the set of
all best response policies to the worst-case prior, and hence the
minimax policy.

\subsection{Gradient descent ascent}
\label{sec:gradient-methods}
We want to calculate the minimax pair $(\pol^*, \bel^*)$ for the Bayesian regret.
This can be done through gradient descent-ascent (GDA) \citep{lin2020gradient}, which alternates performing a gradient step for the prior and performing a gradient step for the policy. We show convergence guarantees for GDA in the finite MDP setting, for certain parametrisations of the policy.
To calculate the minimax solution for the Bayesian regret, we need the gradient with respect to the policy and the prior.
\begin{align}
	\nabla_\pol \eregret(\pol, \bel)
	&= - \int_\MDPs d\bel(\mdp) \nabla_\pol \util(\pol, \mdp)\\
	\nabla_\bel \eregret(\pol, \bel) 
	&= \int_\MDPs \regret(\pol, \mdp) \nabla_\bel d \bel(\mdp).
\end{align}

Intuitively, Algorithm~\ref{alg:gda} works as follows: First, we sample $M$ MDPs from the current prior $\bel_{t-1}$. We use those to do a policy gradient step and obtain a new policy $\pol_{t}$ using standard policy gradient algorithms, as well as a gradient step in the prior space to obtain a new prior $\bel_t$. Since each gradient may not be exact, we use  $G_\pol(\pol,\bel)$ and $G_\bel(\pol,\bel)$ to denote the approximate gradient with respect to the policy and prior respectively.  Appendix~\ref{sec:gradient} describes how we obtain those in detail.
 Since gradient steps may lead us outside the feasible prior space $\mathcal{B}$, we use a projection $\mathcal{P}_{\mathcal{B}}$ to ensure we have a valid prior distribution. Finally, we return a randomly selected policy-prior pair from the ones generated during the algorithm's run.

\begin{algorithm}
	\caption{Stochastic GDA}
	\label{alg:gda}
	\begin{algorithmic}
		\State \textbf{Input}  policy $\pol_0$, belief $\bel_0$, learning rates $(\eta_\pol,\eta_\bel)$ and stochastic gradient estimators $G_\pol,G_\bel$ for $\nabla_\pol \eregret ,\nabla_\bel \eregret$
		\For {$t=1, \ldots, T$}
		\State Get directions $g_\bel = \frac{1}{M}\sum_i G^{(i)}_\bel(\pol_{t-1},\bel_{t-1})$ and $g_\pol = \frac{1}{M}\sum_i G^{(i)}_\pol(\pol_{t-1},\bel_{t-1})$ using M i.i.d samples 
		\State $\pol_t \leftarrow \pol_{t-1} - \eta_\pol g_\pol$
		\State $\bel_t \leftarrow \mathcal{P}_\mathcal{B} \Big(\bel_{t-1} + \eta_\bel g_\bel \Big)$ 
		\EndFor
		\State \textbf{Output} $\bel^*,\pol^*$ uniformly at random from $\{(\bel_1,\pol_1), \ldots ,(\bel_T,\pol_T)\}$
	\end{algorithmic}
\end{algorithm}

\subsubsection{Convergence guarantees for finite set of MDPs}
In the MDP setting with $n$ MDPs, we have $\mathcal{B}$ as the probability simplex which has the diameter $D=\sqrt{2}$. 
Additionally, the gradient 
\begin{align}
	\nabla_\bel \eregret(\pol, \bel) &= \sum_i^n \regret(\pol, \mdp_i) \nabla_\bel P(\mdp_i|\beta) \\
	\nabla_{\bel_i} \eregret(\pol, \bel)& = \regret(\pol, \mdp_i)
\end{align}
is constant and therefore convex. 
\begin{lemma}
	\label{lem:smoothlipschitz}
	If the policy $\pi$ is parameterised as a softmax over actions, independently for each $h_t$ and the horizon $T$ is fixed. Then $\eregret(\pol,\beta)$ is $T^2(|\CA|+1)$-smooth and $\eregret(\cdot,\beta)$ is $T^2$-Lipschitz
\end{lemma}
With these properties, and a batch size $M=1$, the requirements of
Theorem 4.9 of \cite{lin2020gradient} are fulfilled and
Algorithm~\ref{alg:gda} will find a $\epsilon-$stationary point in
terms of Moreau envelopes, given appropriate step sizes, with an
iteration complexity of
\begin{equation}
	\mathcal{O}\left(|\CA|^3T^6\left(\frac{\left(T^4+\sigma^2\right)  \widehat{\Delta}_{\Phi}}{\epsilon^6}+\frac{  \widehat{\Delta}_0}{\epsilon^4}\right) \max \left\{1, \frac{\sigma^2}{\epsilon^2}\right\}\right), 
\end{equation}
as long as
$\mathbb{E}_G\left[\|G(\pol,\bel)-\nabla
  \eregret(\pol,\bel)\|^2\right] \leq \sigma^2.$ Note that no
guarantees exist for general non-convex non-concave Bayesian regret
$\eregret$, as is the case for Dirichlet beliefs and parametric
policies.

Here the stationarity is defined as $||\grad \Phi_{1/2l}(\pi)||_2\le \epsilon$ as in \cite{lin2020gradient}. We have $\Phi(\cdot) = \max_{\bel\in \Bels} \eregret(\cdot, \bel)$ and  $\Phi_\lambda(\pi) = \min_{w \in \Pols} \Phi(w) +(1/2\lambda)||\omega-\pi||_2^2$ is the Moreau envelope of $\Phi$.  Finally we obtain $\widehat{\Delta}_\Phi = \Phi_{1/2l}(\pol_0)- \min_\pol \Phi_{1/2l}(\pol)$ and $\widehat{\Delta}_0 = \Phi(\pol_0) - \eregret(\pi_0, \bel_0)$.

\subsection{Cutting planes}
\label{sec:cutting-planes}

In this section we demonstrate an efficient method for localising the minimax pair $(\pol^*, \beta^*)$ for beliefs over a finite set of MDPs, given that an oracle for the Bayes-optimal policy for a given belief is available. This could for example be obtained in finite horizon tasks with a sufficiently small horizon such that a tree-policy  is tractable. An example of this can be found in \cite[Section 1.5]{duff2002olc}.

\begin{figure}[h]	
		\centering
		\includegraphics[width=0.85\linewidth]{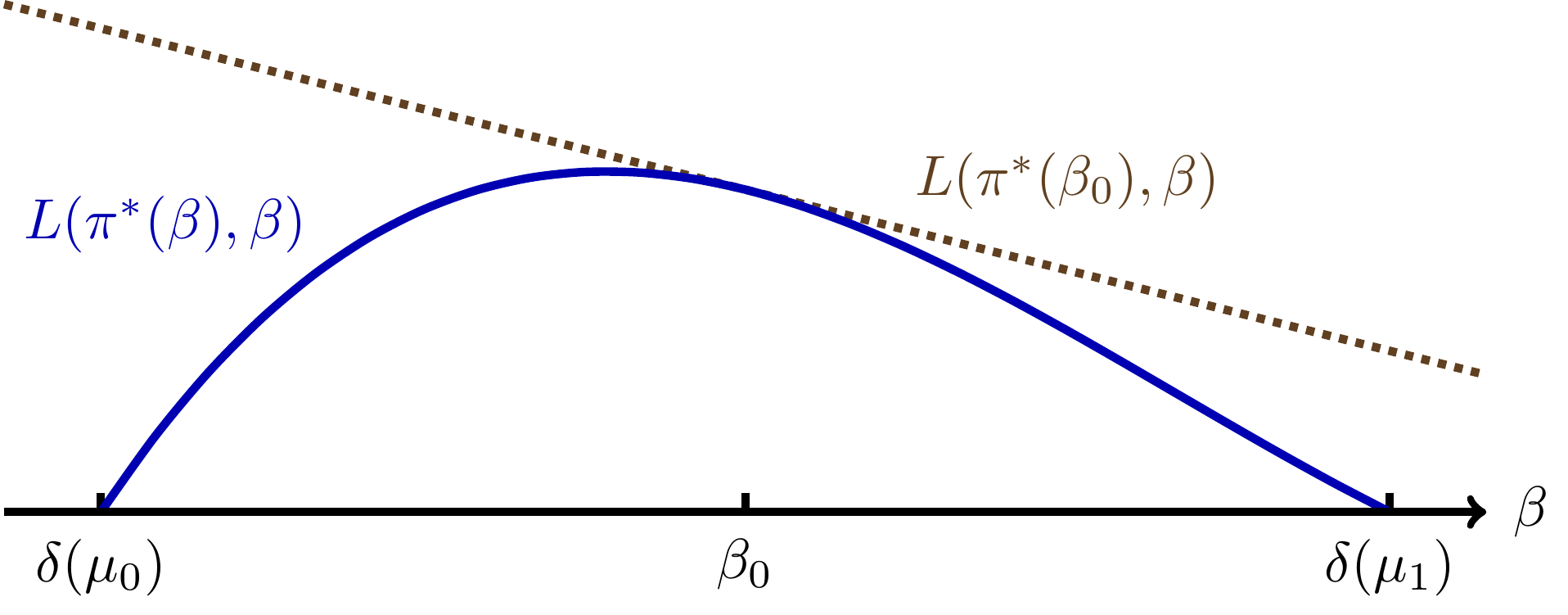}
		\includegraphics[width=0.85\linewidth]{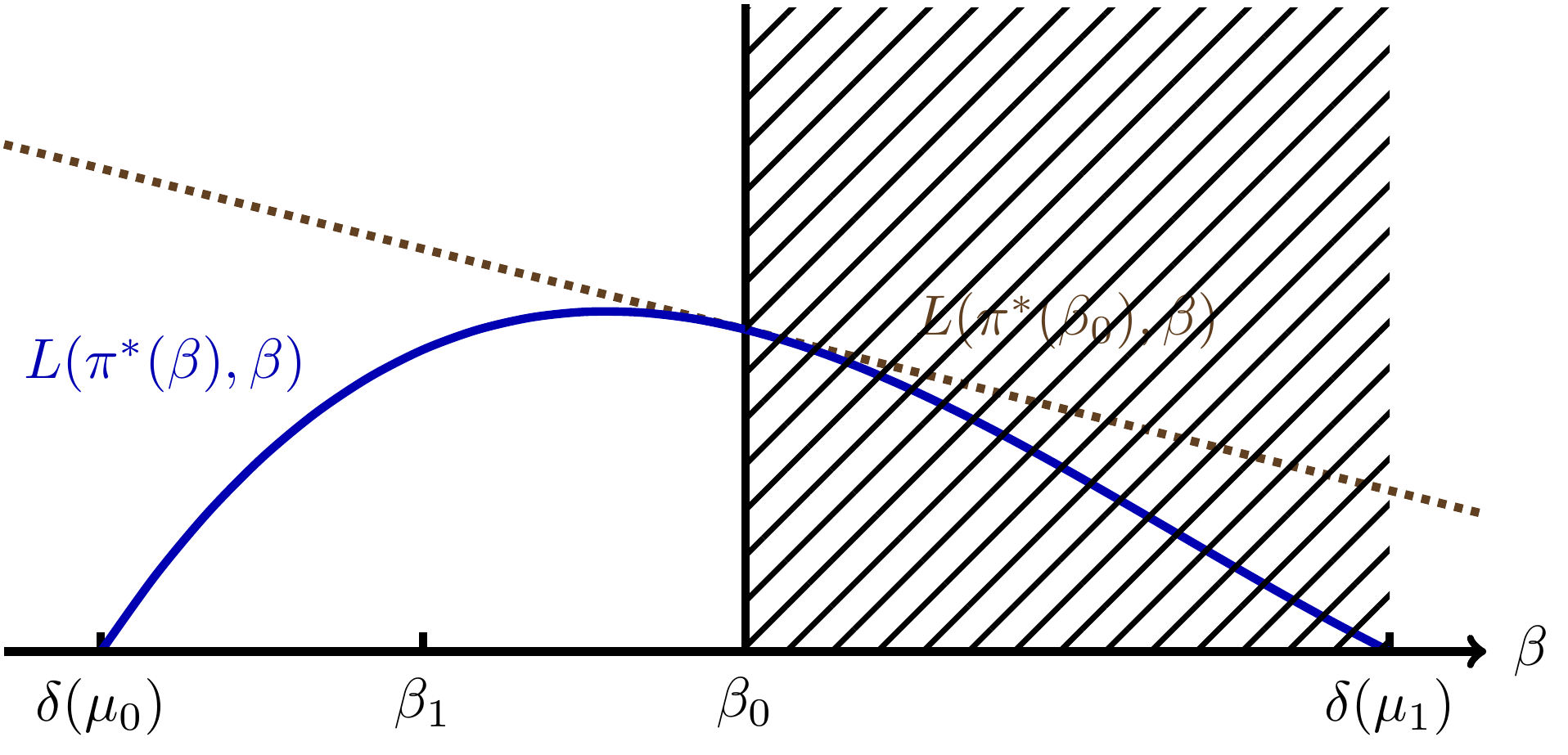}
	\caption{Illustration of cutting plane algorithm for two dimensions. The top image illustrates the Bayesian regret plane obtained for queried belief $\bel_0$ while the bottom image shows how the cut obtained by the plane discards the right side of the belief space and a new queried belief $\bel_1$ is obtained.}
	\label{fig:cutting}
\end{figure}

We use the approximate centroid cutting plane algorithm from \cite{10.1145/1008731.1008733}, which can be seen as a high dimensional extension of the bisection algorithm. The goal here is to find a way to repeatedly obtain a plane where we can reject one side of the half-plane, quickly shrinking the plausible set of beliefs. Each policy $\pol$ has a corresponding regret plane\footnote{Due to the Bayesian regret being an expectation over MDPs and hence is linear.} $\eregret(\pol, \beta)$ over~$\beta$. Since $\eregret(\pol^*(\beta), \beta)) \le \max_{\bel\in \Bels} \eregret(\pol^*(\beta), \beta))$, any $\beta: \eregret(\pol^*(\beta'), \beta'))> \eregret(\pol^*(\beta'), \beta))$ can not be the minimax $\bel$ and can be discarded. This is the same as  discarding the half-plane given by the descent direction of the Bayesian regret plane. An illustration of this principle in two dimensions can be found in Figure \ref{fig:cutting}.

Selecting a new approximate centroid as the next $\beta$ to query guarantees fast convergence in the volume of the plausible set of beliefs given the following lemma. 
\begin{lemma}[Lemma 5 \cite{10.1145/1008731.1008733}]
  Each cut in Algorithm \ref{alg:cutting} will reduce the volume of the set $K_t$ by at least 1/3 with high probability.
\end{lemma}
%The result can be found in Lemma 5 in \cite{10.1145/1008731.1008733}. %\footnote{This generalises \cite{grunbaum1960partitions} which states that cuts centroid reduce the volume with at least $1/e$, unfortunately finding the exact centroid is computationally demanding.}.

The full procedure is described in Algorithm~\ref{alg:cutting}. Here $\beta_t$ is the approximate centroid (through one of the methods in \cite{10.1145/1008731.1008733}, such as hit-and-run sampling) of the set $K_t$. $K_t$ contains the plausible beliefs that could be the minimax belief, at step t of the algorithm. The cut is given by $C_t$ which is the normal to the Bayes regret plane at $\beta_t$ where each element $C_t^{(i)}=\regret(\pol^*(\beta_t), \beta=\delta_{\mu_i})$.

\begin{algorithm}
	\caption{Cutting plane algorithm for finding minimax belief}
		\label{alg:cutting}
	\begin{algorithmic}
		\State \textbf{Input:} Initial belief set of constraints $K_0$, Optimal Policy oracle,  Policy evaluation oracle, $t=0$; 
		\For {$t \in 0, \ldots, T-1$}
		\State Obtain $\beta_t \approx \E_{K_t}[x]$ 
		%\State Calculate $\hat{A}=1/k \sum b_i b_i^T, \hat{B} = \sqrt{\hat{A}^{-1}}, \beta_t = 1/k \sum_i b_i$
		%\State $\epsilon_t = f(|\hat{B}|, \delta)$ \#TBD, take approximate nature of $\hat{B}$ into account.
		\State Obtain optimal policy $\pol^*_{\beta_t}$ and $C_t^{(i)}=\regret(\pol^*(\beta_t), \beta=\delta_{\mu_i})$.
		%\If{$|C_t|\le \delta$}
		%	\State Break
		%\EndIf
		%\State $c_t = C_t/||C_t||$
		\State $K_{t+1} = K_t \cap \{\beta: C_t^T(\beta-\beta_t) > 0\}$
		%\State Optional: Christos extra cut.
		%\State $t = t+1$
		\EndFor
		\State Return $\beta^*\in K_T$ that has $\frac{\text{VOL}(K_T)}{\text{VOL}(K_0)}<\left(\frac{2}{3}\right)^T$ with high probability and corresponding $\pol^*(\beta^*)$.
	\end{algorithmic}
\end{algorithm}

This method is also applicable when the policy space is a set of $\epsilon$-optimal policies $\Pols^\epsilon \subset \Pols$, i.e.\ such that $\max_{\pol \in \Pols^\epsilon} \util(\pol, \bel) \geq \max_{\pol \in \Pols} \util(\pol, \bel) - \epsilon$ for any $\bel \in \Bels$. It is natural to look at such a policy space, because policies obtained through look-ahead tree search or neural network may be adaptive, but they can only be $\epsilon$-optimal in general.
\begin{lemma}
	\label{lemma:approximate_convex}
	If $\max_{\pi \in \PolsEps}\eregret(\pi, \beta) \le \max_{\pi \in \Pols}\eregret(\pi, \beta) + \epsilon$ for all $\bel\in\Bels$ then 
	\begin{align}
		\min_{\pi \in \Pols}\eregret(\pi, \bel^{\epsilon,*})\ge\max_{\bel \in \Bels} \min_{\pi \in \Pols}\eregret(\pi, \bel) - \epsilon
	\end{align} where $\bel^{\epsilon,*} = \argmax_{\bel \in \Bels} \min_{\pi \in \PolsEps}\eregret(\pi, \bel)$.
	
	Additionally, if $\min_{\pi \in \Pols}\eregret(\pi, \bel)$ is c-concave in $\bel$ then $||\bel^{\epsilon,*}- \bel^*||_2<\sqrt{\epsilon/c}$.
\end{lemma} 
A proof is provided in the appendix.

\section{Experiments}
\label{sec:experiments}
We perform three experiments to see how minimax priors differ from
common uniform priors, and examine the relative robustness of the
corresponding policies. The first characterises worst-case priors for
Bernoulli bandits. The second experiment is on finite MDP sets with a
finite horizon. Here we verify the feasibility of the cutting plane
algorithm for finding minimax solutions. We also illustrate the regret
of posterior sampling. The final experiment is for the general
case of discrete MDPs and parametric adaptive policies, where a value
may not exist.\footnote{The code is made available at \url{https://github.com/minimaxBRL/minimax-bayes-rl}.}

\subsection{Illustrations of Worst-Case Priors for Bernoulli Bandits} 
We are interested in analysing the worst-case priors when the Bayesian agent is
responding to nature's prior with a Bayes-optimal policy. In general,
computing the Bayes-optimal policy is intractable. However, for
Bernoulli bandits with infinite horizon and geometrically discounted rewards, so that the utility is defined as
$\sutil = \sum_t \gamma^t r_t$, Gittins
\citep{gittins1979bandit, gittins2011multi} showed that an index
policy, the so-called Gittins index, yields a Bayes-optimal
policy. 

\begin{figure*}[ht!]
	\centering
	\begin{subfigure}{0.245\linewidth}
		\centering
		\includegraphics[width=.8\textwidth]{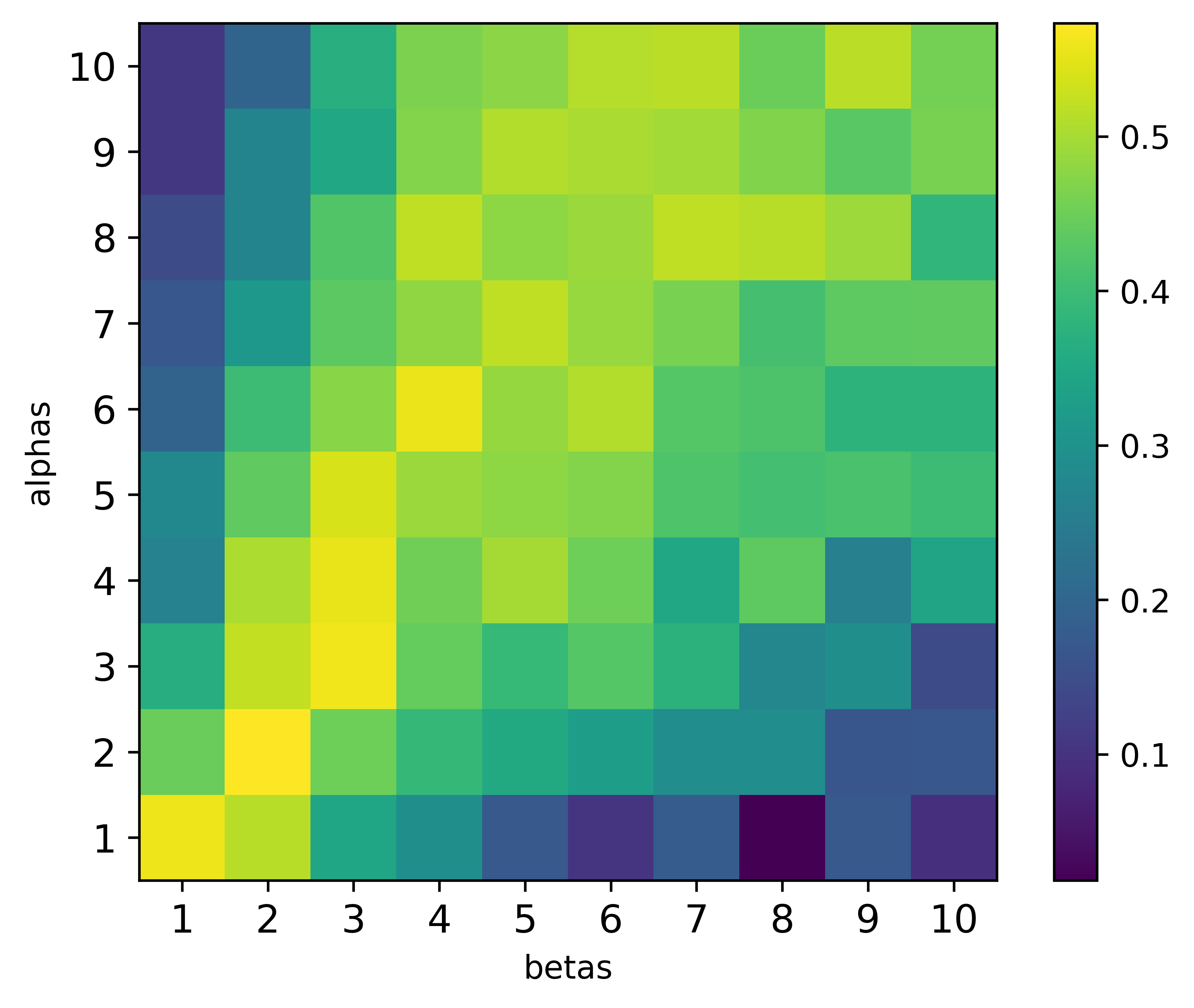}	\caption{$\Beta(1,1)$}
	\end{subfigure}
	%\hspace{-0.05\textwidth}
	\begin{subfigure}{0.245\linewidth}
		\centering
		\includegraphics[width=.8\textwidth]{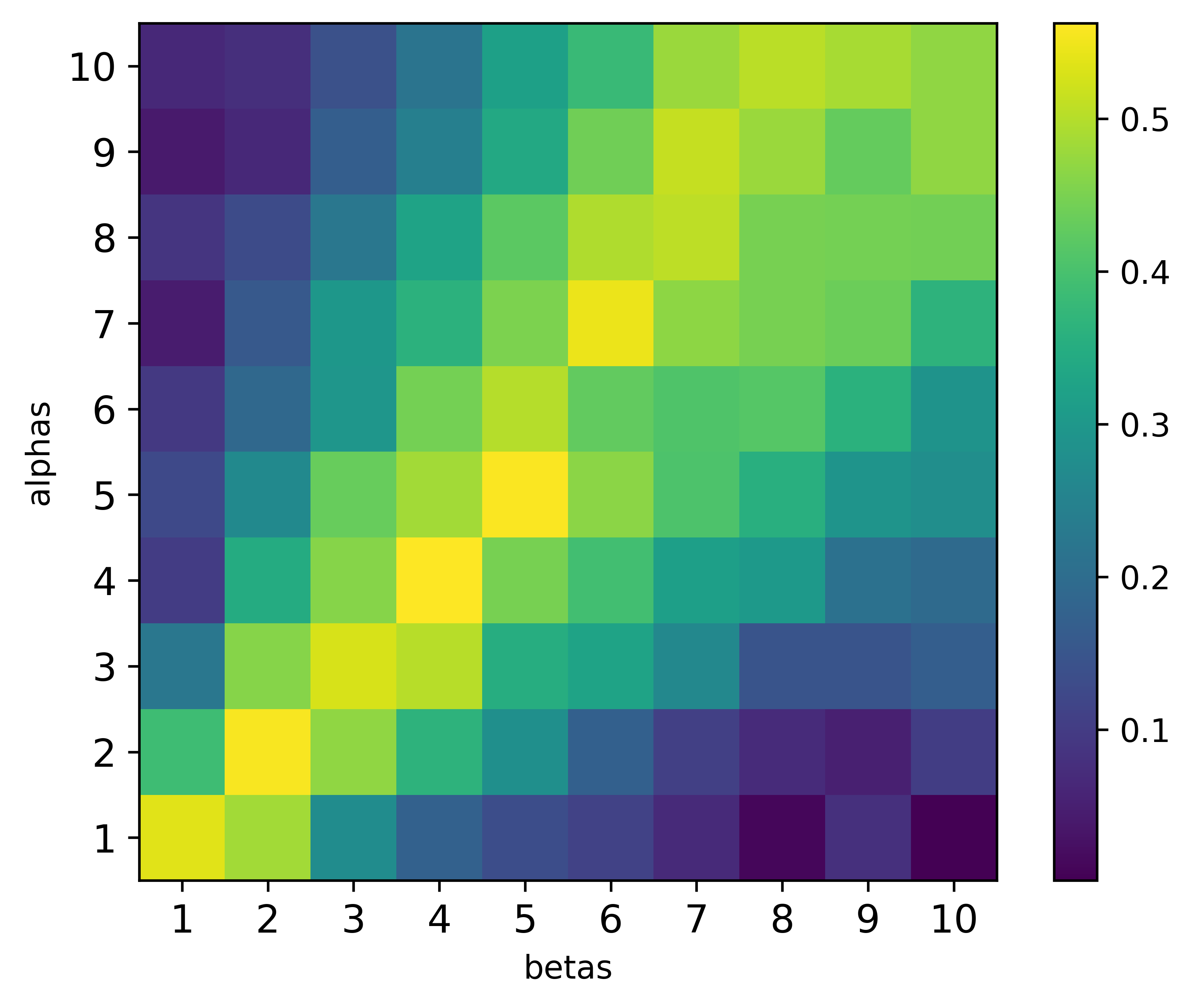}	\caption{$\Beta(3,3)$}
	\end{subfigure}
	\begin{subfigure}{0.245\linewidth}
	\centering
	\includegraphics[width=.8\textwidth]{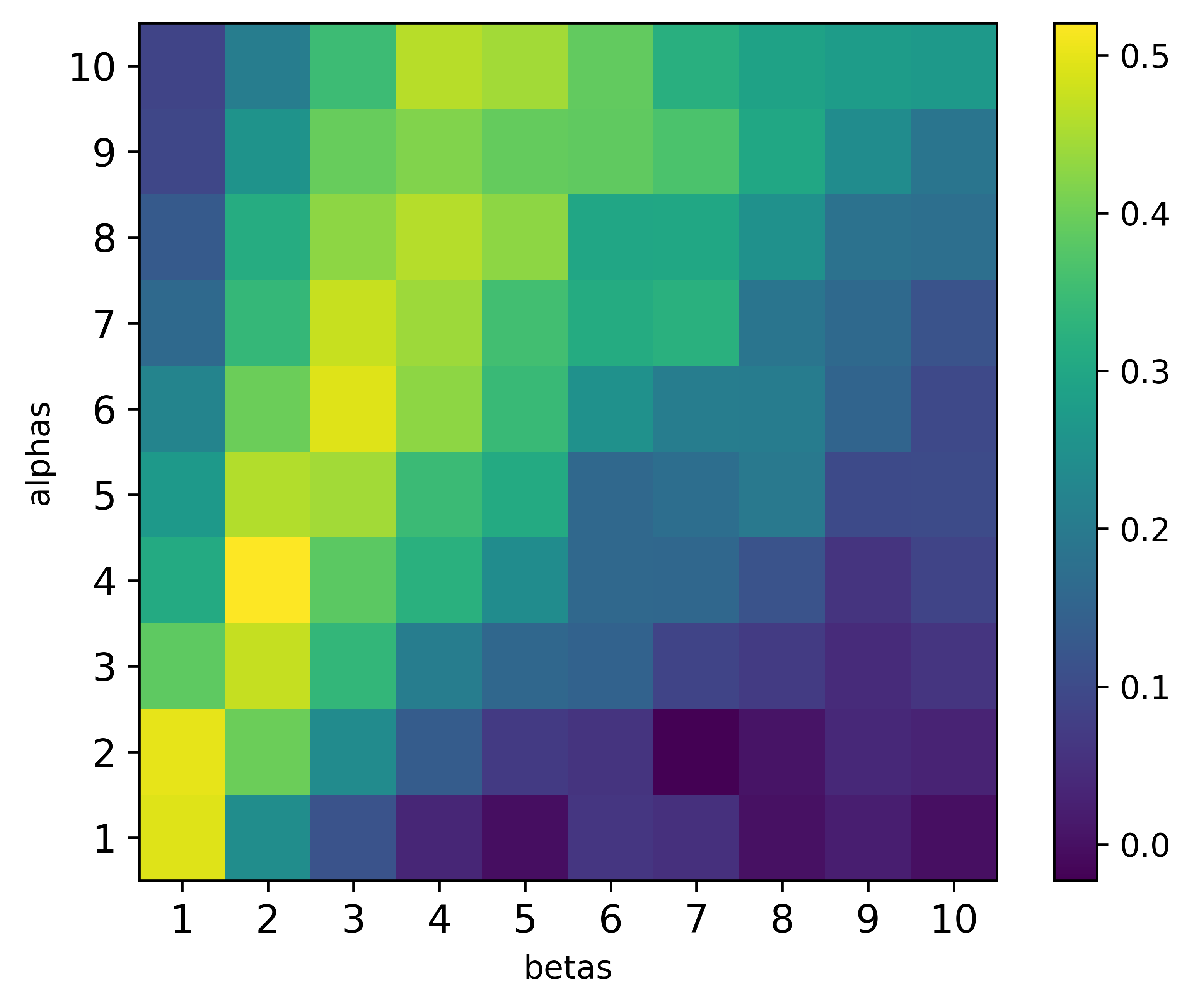}	\caption{$\Beta(4,2)$}
	\end{subfigure}
	%\hspace{-0.05\textwidth}
	\begin{subfigure}{0.245\linewidth}
	\centering
	\includegraphics[width=.8\textwidth]{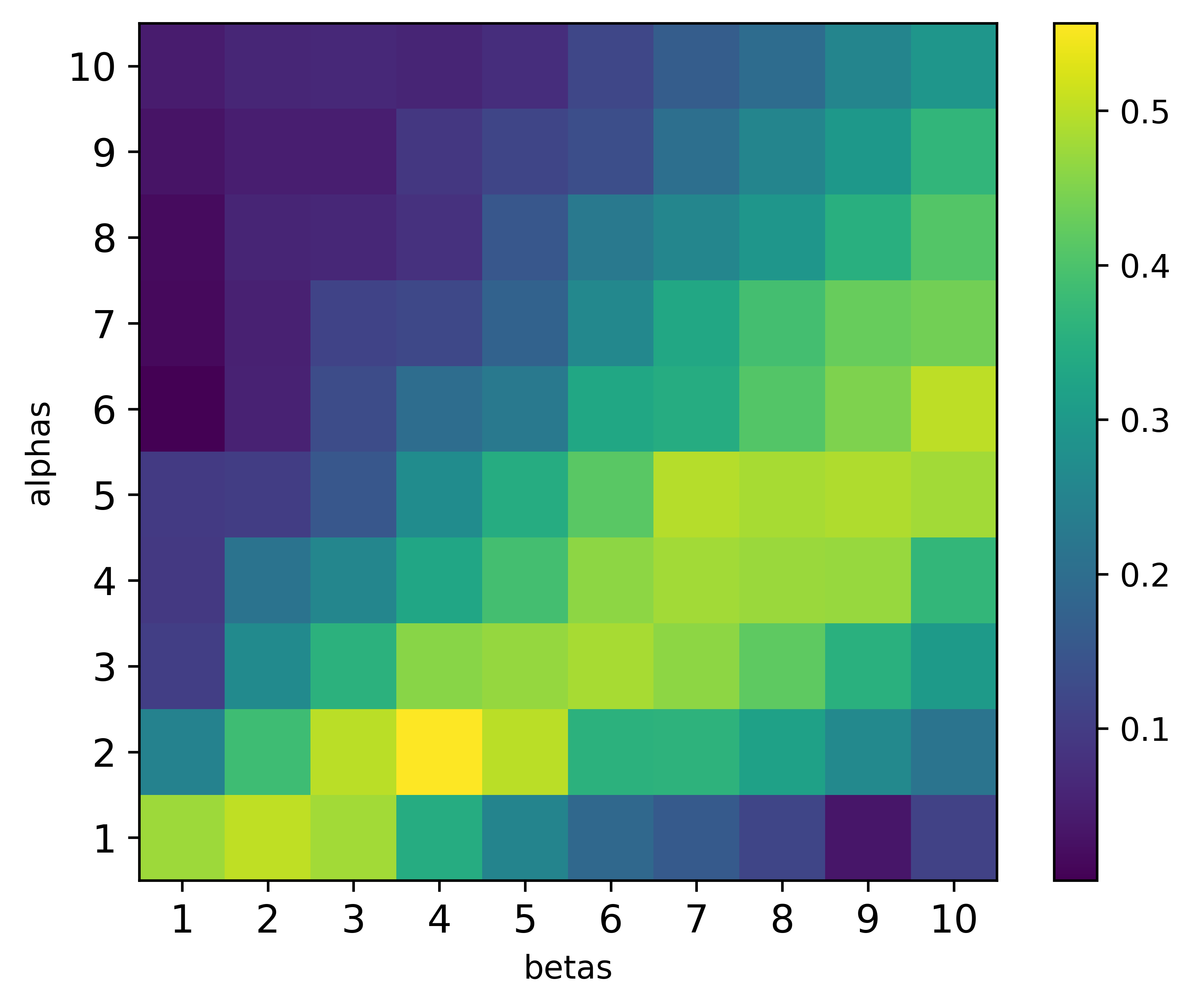}	\caption{$\Beta(2,4)$}
\end{subfigure}
	\caption{The Bayesian regret of the Bayes-optimal policy in two-armed Bernoulli bandits, where the first arm's prior is fixed. The $x$- and $y$-axis denote the parameters of the second arm's prior.}
	\label{fig:gittins_bandits}
\end{figure*}

For $K$-armed Bernoulli bandits
${\theta} = (\theta_1, \dots, \theta_K)$ with $\theta_k \in [0,1]$, we
then consider Beta product priors such that
$\bel({\theta})= \prod_{k=1}^K \Beta(a_k, b_k)\{\theta_k\}$. To
illustrate how the Bayes-expected regret of the Bayes-optimal policy
changes with respect to the prior, we consider a two-armed Bernoulli
bandit, where the first arm's prior is fixed to some distribution
$\Beta(a_1, b_1)$ and the second arm's prior $\Beta(a_2, b_2)$ is set
to different values. Figure~\ref{fig:gittins_bandits} shows the
Bayesian regret for different fixed priors for arm $1$ and varying
prior for arm~$2$.

We observe that high Bayesian regret is typically suffered when the second prior's mean approximately matches the mean of the first arm's prior, i.e.\ $\E[\Beta(a_1, b_1)] = \E [\Beta(a_2, b_2)]$. Moreover, it seems that maximal Bayesian regret is achieved at a completely symmetric prior, i.e.\ $\Beta(a_1, b_1) = \Beta(a_2, b_2)$, irrespective of how the first arm's prior is chosen. More generally, we can observe that lower values of $a$ and $b$ yield higher Bayesian regret, making the intuition precise that the Bayes-optimal policy suffers higher Bayesian regret when the prior provides less information.  
Based on this, a worst-case prior can be conjectured to make arms maximally indistinguishable a priori; as one may expect. 
%
%(The discount factor is set to $\gamma = 0.9$ in the experiments)
% Note that the noise in the plots is due to the approximation of the Gittins index as well as due to the rollouts.)

We also allowed all priors to vary to discover the actual worst-case prior. We found this depends heavily on the discount factor $\gamma$ and the number of arms $K$. For $K=2$ and $\gamma = 0.9$, we found it is approximately $\Beta(0.8, 0.8)$ for both arms. In general, the worst-case prior is symmetric with parameters increasing in the number of arms and the discount factor, i.e.\ moving towards short-tailed priors. 

\subsection{Finite Set of MDPs}
\label{sec:finitemdp}
In this section, we study the properties of minimax problems where we have a belief over a finite set of MDPs. The transition matrix is randomly sampled from an exponential distribution before being normalised. The agent starts in state 1, and the reward is 1 for taking the first action in state N, and zero elsewhere. We use a finite horizon $T=5$ to allow exact computation of the optimal policies and Bayesian regret. Additionally we use $\gamma=1$.

\begin{figure}[ht]

	\centering
	\includegraphics[width=0.95\linewidth]{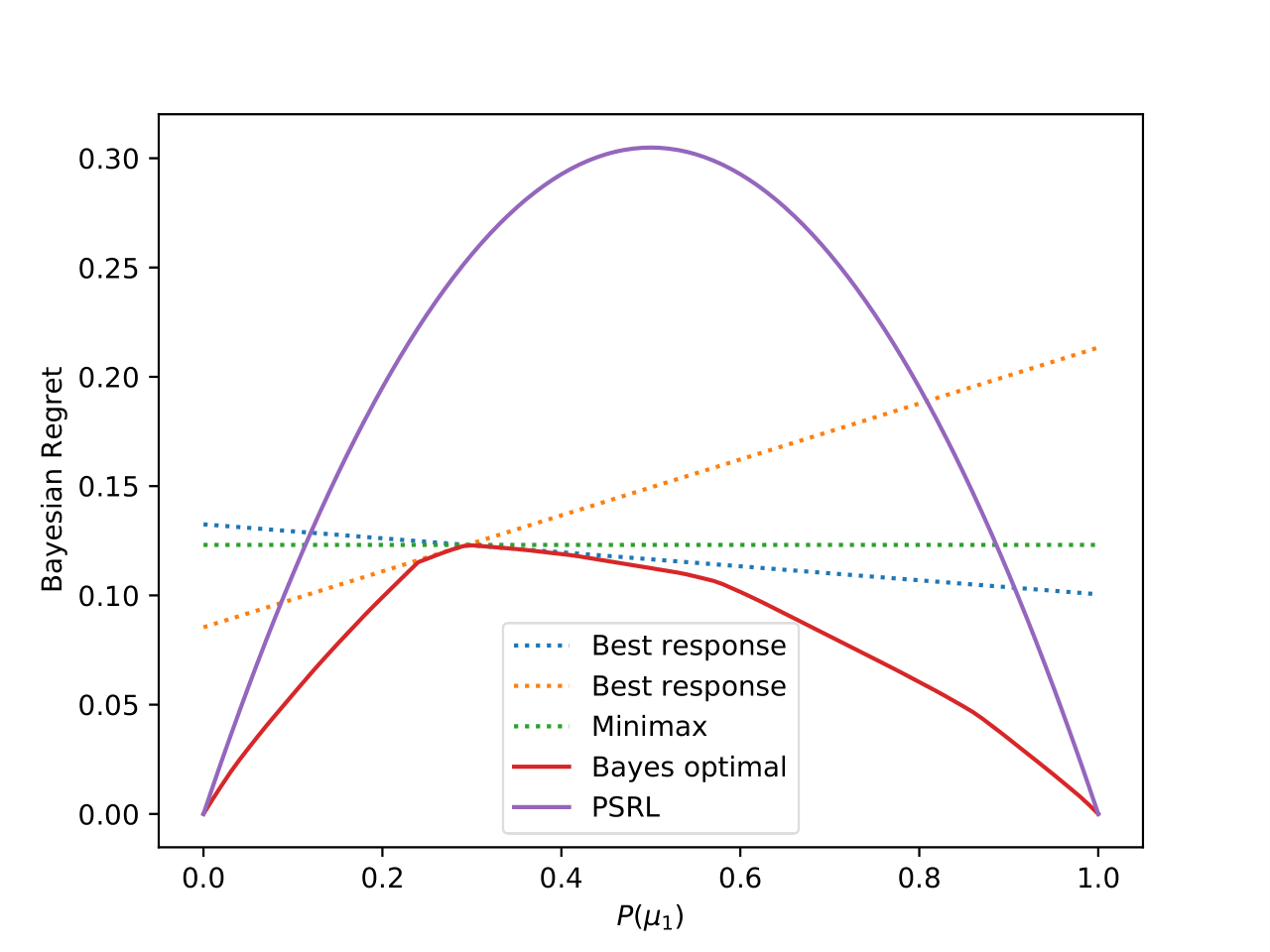}	\caption{This figure shows the Bayesian regret of different policies.  The dashed lines show the value of three adaptive policies optimal for the maximin-regret prior. Two of them are best responses, which are also optimal on either side of the maximin point. The minimax-regret policy is shown in green, and it has a uniform regret no matter what the actual prior is. The solid lines show policies which have knowledge of the MDP prior: the Bayes-optimal policy and the best PSRL policy for that specific prior. Their dependency on the prior makes their regret a concave function.}
	\label{fig:two_mdplineplot}
\end{figure}

Figure \ref{fig:two_mdplineplot} show the Bayesian regret for a two-MDP task. This helps us visualise that the Bayes-optimal value is a piecewise linear function consisting of the minimum over locally optimal policies. We also compare with the Bayesian regret of the PSRL policy~\citep{strens2000bayesian}, which for every episode acts optimally with respect to a sampled MDP from the belief. The quadratic curve for PSRL is due to the fact that we allow the policy to change with the belief.% Note that this is while knot necessarily the same as the gradient of the Bayesian regret with respect to the belief as this would also change the policy, although they are related. 

In additional experiments in Appendix~\ref{appendix:finitemdp}, we study the Bayesian regret landscape for a three MDP setup (see Figure~\ref{fig:mdpgrid}). We also compare the worst case Bayesian regret of the minimax policy and of the Bayes optimal policy for the uniform belief for a few different setups with 16 different MDPs in Table \ref{tab:15mdp} and can see that the minimax policy significantly outperforms the uniform best response policy.

\subsection{Infinite Set of MDPs}

In the following experiments, we study priors over an infinite space of MDPs. The main prior of interest is Dirichlet product-priors. We use the minimax policy gradient algorithm to simultaneously update the parameters of the belief $\bel$ and the parameters of the policy $\pol$. We choose a history-dependent policy parametrisation using a softmax rule. In these experiments we study MDPs with 5 states and two actions. Further, we consider problems with horizon $T = 1000$.

\begin{figure}[ht]
  \centering
	\includegraphics[width=1\linewidth]{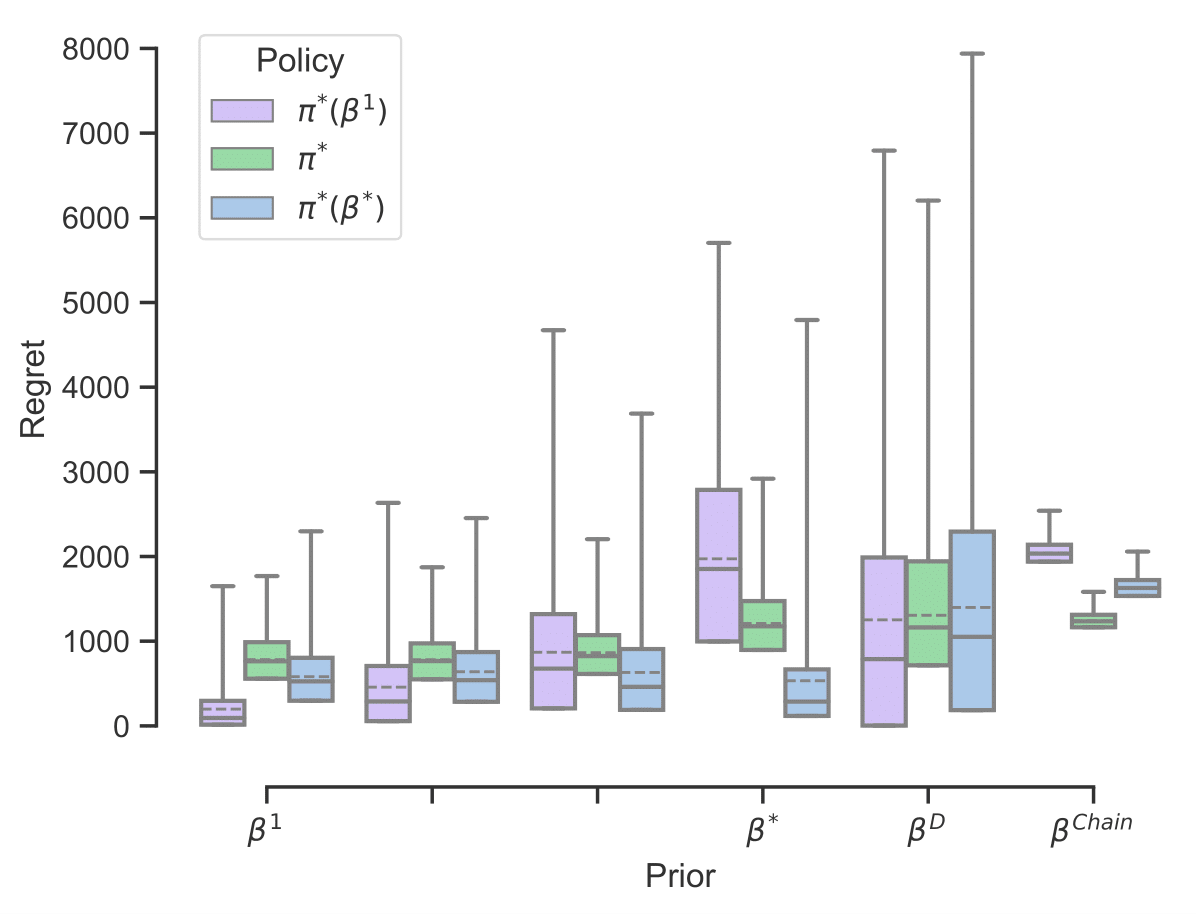}
  \caption{$\beta^D$ is approximately uniform over deterministic MDPs. $\beta^{Chain}$ is a delta distribution over the Chain MDP. The MDPs in between $\beta^1$ (Uniform) and $\beta^*$ (Maximin) are interpolated. The mean is depicted with a dashed line, the solid line is the median and the upper whisker is the $99.9\%$ percentile.}
  \label{fig:boxplot}
\end{figure}
In Figure~\ref{fig:boxplot} we investigate the performance of the minimax policy $\pol^{*}$ compared to the baseline \emph{best response} adaptive policies, $\pol^{*}(\beta^1), \pol^{*}(\beta^*)$, to the uniform prior $\beta^1$ and the maximin prior $\beta^*$, respectively. The three policies are evaluated on six different priors. These are, the uniform prior $\beta^1$, the maximin prior $\beta^*$, two priors interpolated between the uniform and the maximin prior, a uniform prior over deterministic MDPs $\beta^D$ and a delta distribution over the parameters of the Chain environment~\citep{strens2000bayesian}, $\beta^{Chain}$. 

In this setting we can only expect to find approximate minimax solutions. Thus, there is no guarantee the obtained minimax solution is globally robust to changes in belief. However, in Figure~\ref{fig:boxplot} we observe the minimax policy $\pol^{*}$ to be the most robust taking all priors into account.

\section{Discussion and Conclusion}
\label{sec:conclusion}

\paragraph{Related work}
We studied the problem of minimax-Bayes reinforcement learning. Although minimax-Bayes problems are well-known in statistical inference~\cite[c.f.][]{berger:statistical-decision-theory}, they have received little attention in sequential problems. Older work such as \cite{arrow1949bayes} is interested in minimax and Bayes optimal solutions to decision making tasks but without combining them. Similarly, \cite{hodges1952use} relaxes the property of minimax risk to restricted Bayes solutions where the maximal risk is bounded while also changing the objective to an interpolation between the expected and maximal risk. While this is work in the same spirit as ours it is fundamentally different.  \citet{grunwald-dawid:game-robust-bayesian:aos:2004} studied the problem of one-shot experiment design prior to estimation. In the partial monitoring setting,~\citet{lattimore2019information} made connections between the Bayesian minimax regret and the minimax regret.  

There have been a variety of work interested in using meta learning to create Bayes-(adaptive) optimal agents such as \cite{hochreiter2001learning,wang2016learning, mikulik2020meta, zintgraf2021varibad}. They use recurrent neural networks to encode an episode's history so as to adapt optimally in a new episode in a new MDP. As they are interested in optimising for specific MDP distribution, $\bel$ is considered fixed and they solve $\max_\pi \E_{\mdp \sim \bel} \util(\mu, \pi)$ without studying $\beta$'s impact on the utility or regret.

Work on Bayesian robust reinforcement learning~\citep{derman2020bayesian, petrik2019beyond} is related in the manner that they search for policies that are robust against interference from nature. The difference is that they wish to find policies that are good against the worst MDP from the set of MDPs that are plausible with respect to a specific posterior, rather than against an adversarial prior. 
\paragraph{Conclusion}

In this work we study the computation of minimax-Bayes policies, which have not been previously considered. We also include conditions for when the solutions can be guaranteed to be found efficiently. Experimentally we find that these policies not only appear to be feasible, but also that such policies can be significantly more robust than those based on standard uninformative priors. Finally, we make exposition of many important properties of minimax-Bayes solutions for reinforcement learning to make a basis for future work in this area.

\section*{Acknowledgements}
	This work was partially supported by the Wallenberg AI, Autonomous Systems and Software Program (WASP) funded by the Knut and Alice Wallenberg Foundation, the Swedish research council grant on ``Learning, Privacy and the Limits of Computation'' and the Norwegian research council grant on ``Algorithms and Models for Socially Beneficial AI''. We are grateful for their support. Many thanks to Emmanouel Androulakis, whose Master thesis developed MWA algorithms for this problem, and to Tor Lattimore for discussions about minimax properties in the Bayesian setting.

\vskip 0.2in
\bibliography{references}
\onecolumn
\appendix
\section{Gradient calculations.}
\label{sec:gradient}
For solving the minimax problem either for the expected utility or the expected regret, we need to calculate the appropriate gradient for both the policy and the prior.
The gradients for the expected utility are as follows:
\begin{align}
  \nabla_\pol \util(\pol, \bel)
  &= \int_\MDPs d\bel(\mdp) \nabla_\pol \util(\pol, \mdp),
  &\nabla_\bel \util(\pol, \bel) 
  &= \int_\MDPs \util(\pol, \mdp) \nabla_\bel d \bel(\mdp)
\end{align}
The Bayesian regret gradient is similarly obtained:
\begin{align}
  \nabla_\pol \eregret(\pol, \bel)
  &= - \int_\MDPs d\bel(\mdp) \nabla_\pol \regret(\pol, \mdp)
  &
   \nabla_\bel \eregret(\pol, \bel) 
&= \int_\MDPs \regret(\pol, \mdp) \nabla_\bel d \bel(\mdp).
\end{align}
Since in the minimax regret scenario, the agent is minimising rather than maximising, the policy update is identical. However, the prior gradient is scaled with respect to the regret rather than the utility. Let us now look at how to calculate those gradients in more detail.

\subsection{Policy gradient}
Here we look at two classes of policies. The first occurs when there
is a finite number of bases (possibly stochastic and behavioural)
policies from which the agent chooses one randomly. The second is a
class of parametrised stochastic behavioural policies. 

\paragraph{Finite policy distributions.}
For a strategy $\str = (\str_1, \ldots, \str_n)$ over a finite set of $n$ policies $\Pols \subset \PolsHS$, we can write
\begin{equation}
   \util(\str, \bel) = \sum_{\pol, \mdp} \str(\pol) \util(\pol, \mdp) \bel(\mdp).
\end{equation}
We then obtain
\begin{equation}
  \p{\str_i} \util(\str, \bel) =  \sum_{\mdp} \util(\pol_i, \mdp) \bel(\mdp)
\end{equation}
We do not use this setting in practice in the paper, but it is an interesting special case.

\paragraph{Stochastic policies.}
Stochastic policies $\pol$ in a parametrised policy space $\Pols_W \subset \PolsHS$ can be an arbitrary neural network policy. For a finite set of MDPs, the gradient is:
\begin{equation}
  \nabla_\pol \util(\pol, \bel) = \sum_{\mdp} \nabla_\pol \util(\pol, \mdp) \bel(\mdp).
\end{equation}
For an infinite set of MDPs, we have
\begin{equation}
  \nabla_\pol \util(\pol, \bel)
  =
  \int_{\MDPs} \nabla_\pol \util(\pol, \mdp) \dd \bel(\mdp)
  \approx
  \frac{1}{M}
  \sum_{k=1}^M \nabla_\pol \util(\pol, \mdp^{(k)}), \, \mdp_k \sim \beta(\mdp)
 \end{equation}

So it is only necessary to compute
\begin{align*}
  \nabla_\pol \util(\pol, \mdp)
  &=
  \sum_h U(h) \Pr_\mdp^\pol(h) \sum_t \nabla_\pol \ln \pol(a_t \mid h_t)\\
  &=
  \sum_h U(h) \Pr_\mdp^\pol(h) \sum_t \frac{\nabla_\pol \pol(a_t \mid h_t)}{\pol(a_t \mid h_t)},
\end{align*}
where for a given history $h = (s_1, r_1, a_1, \ldots, s_T, r_T)$, $h_t =  (s_1, r_1, a_1, \ldots, s_t, r_t)$.
It remains to compute $\nabla_\pol \pol(a_t \mid h_t)$, which can be done automatically using auto-grad software. 

However, one particular case is when the policy is parametrised with $\bw_a = (w_{a,i})_{i=1}^n$ vectors combined with a statistic $\phi : \CH \to \Reals_+^n$ so that
\begin{equation}
  \pol(a_t = a \mid h_t)
  = \frac{\trans{\bw_a} \phi(h_t)}{\sum_b \trans{\bw_b} \phi(h_t)}
  = \frac{\sum_i{w_{a,i} \phi_i(h_t)}}{\sum_b \sum_i w_{b,i} \phi_i(h_t)}
\end{equation}
\begin{equation}
  \p{w_{a,i}} \pol_\bw(a_t = a \mid h_t) = \frac{\phi_{i}(h_t) [\sum_{(b,j) \neq (a,i)} w_{b,j} \phi_j(h_t)]  }{[\sum_b \sum_j w_{b,j} \phi_j(h_t)]^2},
  \qquad
  \p{w_{b,i}} \pol_\bw(a_t = a \mid h_t) = -\frac{\phi_i(h_t) \sum_j w_{a,j}\phi_j(h_t)}{[\sum_b \sum_j w_{b,j} \phi_j(h_t)]^2}.
\end{equation}
With a feature representation $\phi: \CH \times \CA \to \Reals^n$ and
a softmax policy then
\begin{align}
  \pol(a_t  \mid h_t)
  &=
    \frac{e^{\trans{\bw} \phi(h_t, a_t)}}{\sum_b e^{\trans{\bw} \phi(h_t, b)}},
  &
  \nabla_\bw \ln \pol(a_t \mid h_t)
  &=   \phi(h_t, a_t) - \sum_{a \in \CA} \pol(a_t = a \mid h_t) \phi(h_t, a).
\end{align}
For the case where $\phi(h_t, a)$ simply partitions the history, so that $\trans{\bw}\phi(h, a) = w_{h,a}$, the
above becomes
\begin{equation}
	\label{eq:policy_gradient_partition}
  \p{w_{h,a}} \ln \pol(a_t \mid h_t)
  =
  \begin{cases}
    1 - \pol(a | h), & a_t = a, h_t = h\\

    - \pol(a | h), & a_t \neq a, h_t = h\\
    0, & h_t \neq h
  \end{cases}
\end{equation}

\subsection{Prior gradient.}
The steps above were all standard policy gradient steps, which can be implemented with sampled MDPs from the current prior. However, we also need to update the prior distribution with a gradient step. Here we distinguish two cases: a belief over a finite number of MPDs and a Dirichlet belief.

\paragraph{Finite $\CM$.}
Now let us represent the belief as a finite-dimensional vector $\bel = (\bel_i)$ on the simplex. The partial derivative is then:
\begin{equation}
\frac{\partial}{\partial \bel_i}\util(\pol, \bel)
= \sum_j \util(\pol, \mdp_j) \frac{\partial}{\partial \bel_i} \bel_j
= \util(\pol, \mdp_j) 
\end{equation}
\paragraph{Dirichlet $\CM$.}
Let us first consider the general case of an infinite MDP space. Then we can approximate the gradient of the expected utility through sampling:
\begin{equation}
\nabla_\bel \util(\pol, \bel) 
= \int_\MDPs \util(\pol, \mdp) \nabla_\bel \ln[\bel(\mdp)] d \bel(\mdp) 
\approx \frac{1}{M} \sum_{k=1}^M \util(\pol, \mdp^{(k)}) \nabla_\bel \ln[\bel(\mdp^{(k)})],
\end{equation}
where $\mdp^{(k)} \sim \bel$ are samples from the current prior.

For discrete state-action MDPs for a certain number of states and
actions, we can use a Dirichlet-product distribution. This means that for each state-action's $(s,a)$ transition distribution, we define a separate Dirichlet distribution $\bel(\mdp_{s,a})$ with parameter vector $\alpha_{s,a} \in \Reals_+^{|S|}$:
\begin{equation}
\bel(\mdp) = \prod_{(s,a)} \bel(\mdp_{s,a})
,
\qquad
\bel(\mdp_{s,a}) = \frac{1}{B(\alpha_{s,a})} \prod_i \mdp_{s,a,i}^{\alpha_{s,a,i}-1}
\end{equation}
where $\mdp_{s,a,i} = \Pr(s_{t+1} = i | s_t = s, a_t = a)$.
For the sequel, it is notationally convenient to ignore the $s,a$ subscript and focus only on the next state distribution $i$
\begin{align*}
\frac{\partial}{\partial \alpha_j} \ln \bel(\mdp)
&=
\frac{\partial}{\partial \alpha_j} 
\ln \left\{\frac{1}{B(\alpha)} \prod_i \mdp_i^{\alpha_i - 1}\right\}\\
&=
\frac{\partial}{\partial \alpha_j} 
\left\{
\ln \frac{1}{B(\alpha)}
+\sum_i (\alpha_i - 1) \ln \mdp_i
\right\}\\
&=
\frac{\partial}{\partial \alpha_j} 
\ln \frac{1}{B(\alpha)}
+\ln \mdp_j
\end{align*}
Note that
\begin{align*}
\ln 1/B(\alpha) 
&= 
\ln \frac{\Gamma(\sum_i \alpha_i)}{\prod_i \Gamma(\alpha_i)}\\
&=
\ln \Gamma(\sum_i \alpha_i) - \sum_i \log \Gamma(\alpha_i)
\end{align*}
So that
\begin{align*}
\frac{\partial}{\partial \alpha_j} \ln 1/B(\alpha) 
&=
\frac{\partial}{\partial \alpha_j} \ln \Gamma(\sum_i \alpha_i) - \frac{\partial}{\partial \alpha_j} \ln \Gamma(\alpha_j)\\
&=
\frac{1}{\Gamma(\sum_i \alpha_i)} \frac{\partial}{\partial \alpha_j}  \Gamma(\sum_i \alpha_i) - \frac{1}{\Gamma(\alpha_j)} \frac{\partial}{\partial \alpha_j} \Gamma(\alpha_j)\\
&=
\psi(\sum_i \alpha_i) - \psi(\alpha_j)
\end{align*}
where $\psi$ is the digamma function.

This means that the overall derivative is 
\begin{align*}
\frac{\partial}{\partial \mdp_{s,a,i}} \ln \bel(\mdp)
&= 
\frac{\partial}{\partial \mdp_{s,a,i}}
\ln \prod_{(s',a')} \bel(\mdp_{s',a'})\\
&= 
\frac{\partial}{\partial \mdp_{s,a,i}}
\sum_{s',a'} \ln \bel(\mdp_{s',a'})\\
&= 
\frac{\partial}{\partial \mdp_{s,a,i}}
\ln \bel(\mdp_{s,a})\\
&=\psi(\sum_j \alpha_{s,a,j}) - \psi(\alpha_{s,a,i}) + \ln(\mdp_{s,a,i})
\end{align*}
Combinging the above, we get
\begin{equation}
\alpha_{s,a,i}^{(k)}
= \alpha_{s,a,i}^{(k-1)} - 
 \delta^{(k)}\util(\pol, \mdp^{(k)}) \left[\psi(\sum_j \alpha_{s,a,j}) - \psi(\alpha_{s,a,i}) + \ln(\mdp^{(k)}_{s,a,i})\right],
\end{equation}
where $\delta^{(k)}$ is the step-size.

\paragraph{Reward prior.} We can derive a similar update for Beta-distributed rewards, with
\begin{align}
\alpha_{s}^{(k)}
&= \alpha_{s}^{(k-1)} - 
 \delta^{(k)}\util(\pol, \mdp^{(k)}) \left[\psi(\alpha_s + \beta_s) - \psi(\alpha_s) + \ln(\rho^{(k)}_{s})\right]\\
\beta_{s}^{(k)}
&= \beta_{s}^{(k-1)} - 
 \delta^{(k)}\util(\pol, \mdp^{(k)}) \left[\psi(\alpha_s + \beta_s) - \psi(\beta_s) + \ln(1-\rho^{(k)}_{s})\right].
\end{align}

We can also define the Beta-distribution with alternate parametrisation: $p_s=\alpha_s/(\alpha_s+\beta_s),\, n_s=\alpha_s+\beta_s$ which implies $\alpha_s=p_s n_s, \beta_s=n_s(1-p_s).$
We then obtain
\begin{align}
	\frac{\partial}{\partial p_s} \ln \bel(\mdp) \\
	& = n_s \frac{\partial}{\partial \alpha_s} \ln \bel(\mdp) - n_s\frac{\partial}{\partial \beta_s} \ln \bel(\mdp) \\ 
	& = n_s\left[-\psi(\alpha_s)+\psi(\beta_s) + \ln(\rho^{(k)}_{s}) - \ln(1-\rho^{(k)}_{s})\right] \\
	& = n_s\left[-\psi(\alpha_s)+ \psi(\beta_s) + \ln\left(\frac{\rho^{(k)}_{s}}{1-\rho^{(k)}_{s}}\right)\right]
\end{align}

\begin{align}
	\frac{\partial}{\partial n_s} \ln \bel(\mdp) 
	& = p \frac{\partial}{\partial \alpha_s} \ln \bel(\mdp) + (1-p)\frac{\partial}{\partial \beta_s} \ln \bel(\mdp) \\ 
	& = p\left[-\psi(\alpha_s)+ \psi(\beta_s) + \ln(\rho^{(k)}_{s}) - \ln(1-\rho^{(k)}_{s})\right] +  \left[\psi(\alpha_s + \beta_s) - \psi(\beta_s) + \ln(1-\rho^{(k)}_{s})\right] \\
	& = p\left[-\psi(\alpha_s)+ \psi(\beta_s) + \ln\left(\frac{\rho^{(k)}_{s}}{1-\rho^{(k)}_{s}}\right)\right] + \left[\psi(\alpha_s + \beta_s) - \psi(\beta_s) + \ln(1-\rho^{(k)}_{s})\right]
\end{align}

\section{Ommitted proofs}
\label{sec:ommitted-proofs}
\begin{proof}[Proof of Lemma~\ref{lem:tor-pc}.]
	For any $\bel$
	\begin{align}
		\max_{\mdp \in \MDPs} \regret(\pol, \mdp)
		& \geq \max_{\mdp \in \supp(\bel)} \regret(\pol, \mdp)\\
		& = \max_{\mdp \in \supp(\bel)} \util(\pol^*(\mdp), \mdp) - \util(\pol, \mdp)\\
		& \geq \max_{\mdp \in \supp(\bel)} \util(\pol^*(\bel), \mdp) - \util(\pol, \mdp)\\
		& \geq \sum_{\mdp \in \supp(\bel)} \bel(\mdp) [\util(\pol^*(\bel), \mdp) - \util(\pol, \mdp)]\\
		& =\util(\pol^*(\bel), \bel) - \util(\pol, \bel)
		=\regret(\pol, \bel).
	\end{align}
	Since the above holds for any $\bel$, $\max_\mdp \regret(\pol, \mdp) \geq \max_\bel \regret(\pol, \bel)$.
	Letting $\delta(\CM)$ denote the degenerate distributions on individual members of $\CM$, we have:
	\begin{align*}
		\max_\bel \regret(\pol, \bel)
		&\geq
		\max_{\bel \in \delta(\MDPs)} \regret(\pol, \mdp) = 
		\max_{\mdp \in \MDPs} \regret(\pol, \mdp) 
	\end{align*}
\end{proof}

\begin{proof}[Proof of Lemma~\ref{lem:smoothlipschitz}.]

Let $\pi, \pi', \pi''\in \Pols .$
To verify that $\eregret(\pol,\bel)$ is $l$-smooth we study if
\begin{align}
	||\nabla \eregret(\pol, \bel) - \nabla \eregret(\pol', \bel')|| \le l ||(\pol,\beta)-(\pol', \beta')||.
\end{align}

\begin{align}
	&	||\nabla \eregret(\pol'', \bel'') - \nabla \eregret(\pol', \bel')||_2^2 \\
		\le &||(\pol'',\bel'')- (\pol',\beta')||_2^2(\sup_{\pi, \beta} ||\nabla^2 \eregret(\pol, \bel)||_2^2 ) \\
		= & ||(\pol'',\bel'')- (\pol',\beta')||_2^2(\sup_{\pi, \beta} ||\nabla^2_{\pol} \eregret(\pol, \bel)||_2^2) \\
		\le&  ||(\pol'',\bel'')- (\pol',\beta')||_2^2(\sup_{\pi, \beta} ||\nabla^2_{\pol} \eregret(\pol, \bel)||_F^2 )
\end{align}
Here the second transformation is due to the fact that any derivative with respect to $\bel$ is constant, and therefore the second order derivatives are zero except for $\nabla^2_{\pol}$. $||.||_F$ denotes the Frobenius norm.

For stochastic policies $\pol$ in a parametrised policy space $\Pols_W \subset \PolsHS$ , we can write (cf. \cite{dimitrakakis2018decision}):
\begin{equation}
\nabla_{\pol} \eregret(\pol, \bel) = \nabla_{\pol} \util(\pol, \bel) = \sum_{\bel} \nabla_{\pol} \util(\pol, \mdp) \bel(\mdp).
\end{equation}
Similarly, we obtain, for the Hessian:
\begin{equation}
\nabla^2_{\pol} \eregret(\pol, \bel) = \nabla^2_{\pol} \util(\pol, \bel) = \sum_{\bel} \nabla^2_{\pol} \util(\pol, \mdp) \bel(\mdp).
\end{equation}
So it is only necessary to compute

\begin{align}
	\nabla^2_{\pol} \util(\pol, \mdp)
	&=
	\sum_h \util(h) \nabla_\pol(\Pr_\mdp^\pol(h) \sum_t \nabla_\pol \ln \pol(a_t \mid h_t)) \\
	& = \sum_h \util(h) (\nabla_\pol(\Pr_\mdp^\pol(h)) \sum_t \nabla_\pol \ln \pol(a_t \mid h_t)) + \Pr_\mdp^\pol(h) \sum_t \nabla^2_{\pol} \ln \pol(a_t \mid h_t)) \\
	& = \sum_h \util(h) (\Pr_\mdp^\pol(h) \sum_t \nabla_\pol \ln \pol(a_t \mid h_t) \sum_t \nabla_\pol \ln \pol(a_t \mid h_t)^T + \Pr_\mdp^\pol(h) \sum_t \nabla^2_{\pol} \ln \pol(a_t \mid h_t)) \\
	& = \sum_h \util(h) (\Pr_\mdp^\pol(h) \sum_t \nabla_\pol \ln \pol(a_t \mid h_t) \nabla_\pol \ln \pol(a_t \mid h_t)^T + \Pr_\mdp^\pol(h) \sum_t \nabla^2_{\pol} \ln \pol(a_t \mid h_t)) 
\end{align}
where for a given history $h = (s_1, r_1, a_1, \ldots, s_T, r_T)$, $h_t =  (s_1, r_1, a_1, \ldots, s_t, r_t)$.

\iffalse
If we use a feature representation $\phi: \CH \times \CA \to \Reals^n$ and
a softmax policy then
\begin{align}
	\pol(a_t  \mid h_t)
	&=
	\frac{e^{\trans{\bw} \phi(h_t, a_t)}}{\sum_b e^{\trans{\bw} \phi(h_t, b)}},
	&
	\nabla_\bw \ln \pol(a_t \mid h_t)
	&=   \phi(h_t, a_t) - \sum_{a \in \CA} \pol(a_t = a \mid h_t) \phi(h_t, a).
\end{align}
For the case where $\phi(h_t, a)$ simply partitions the history, so that $\trans{\bw}\phi(h, a) = w_{h,a}$, the
above becomes
\fi

From the setting of a softmax policy and a partitioned history in Eq \eqref{eq:policy_gradient_partition}.
\begin{align}
	\label{eq:pol_grad}
\p{w_{h,a}} \ln \pol(a_t \mid h_t)
=
\begin{cases}
	1 - \pol(a | h), & a_t = a, h_t = h\\
	- \pol(a | h), & a_t \neq a, h_t = h\\
	0, & h_t \neq h
\end{cases}\\
	\label{eq:pol_hess}
\frac{\partial\partial}{\partial w_{h,a} \partial w_{h,a'}} \ln \pol(a_t \mid h_t)
=
\begin{cases}
	 \pol(a | h) (\pol(a | h)-1), &  a=a', h_t = h\\
	 \pol(a | h) \pol(a' | h), & a \neq a', h_t = h\\
	0, & h_t \neq h.
\end{cases}
\end{align}

We then get 
Let $\nabla^2_{\pol} \util(\pol, \mdp) = G_1+G_2$ where 
\begin{align}
G_1 &= \sum_h \util(h) \Pr_\mdp^\pol(h) \sum_t \nabla_\pol \ln \pol(a_t \mid h_t) \nabla_\pol \ln \pol(a_t \mid h_t)^T  \\
G_2&=\sum_h \util(h) \Pr_\mdp^\pol(h) \sum_t \nabla^2_{\pol} \ln \pol(a_t \mid h_t).
\end{align}

	\begin{align}
		  ||G_1||_F = &||\sum_h \util(h) \Pr_\mdp^\pol(h) \sum_t \nabla_\pol \ln \pol(a_t \mid h_t) \nabla_\pol \ln \pol(a_t \mid h_t)^T ||_F\\ 
		  \le & \max_h |\util(h)||| \sum_h \Pr_\mdp^\pol(h) \sum_t \nabla_\pol \ln \pol(a_t \mid h_t) \nabla_\pol \ln \pol(a_t \mid h_t)^T||_F  \\
		  \le & T ||\sum_h \Pr_\mdp^\pol(h) \sum_t \nabla_\pol \ln \pol(a_t \mid h_t) \nabla_\pol \ln \pol(a_t \mid h_t)^T||_F  \\
		  \label{eq:g1_pht}
		= & T \sqrt{ \sum_{h_t}  \sum_{a \in \mathcal{A}} \sum_{a' \in \mathcal{A}} \left (\Pr_\mdp^\pol(h_t)T \frac{\partial \ln \pol(a_t \mid h_t)}{\partial \omega_{h_t,a}} \frac{\partial \ln \pol(a_t \mid h_t)}{\partial \omega_{h_t,a'}}\right)^2}  \\
		\label{eq:g1_grad}
		\le &T  \sqrt{ \sum_{h_t} T^2\Pr_\mdp^\pol(h_t)^2 \sum_{a \in \mathcal{A}} \sum_{a' \in \mathcal{A}}  1^2}  \\
		\le &T  \sqrt{ T^2 \sum_{h_t} \Pr_\mdp^\pol(h_t) \sum_{a \in \mathcal{A}} \sum_{a' \in \mathcal{A}} 1}  \\
		\le & T  \sqrt{ T^2 |\mathcal{A}|^2}  \\	
		\le & |\mathcal{A}| T^2
	\end{align}
Here equation~\eqref{eq:g1_pht} comes from the definition of the Frobenius norm and the fact that every element $(h_t,a, a')$ in the matrix corresponds to   $ \sum_{h} \mathbb{I}_{h_t\in h} \Pr_\mdp^\pol(h) \frac{\partial \ln \pol(a_t \mid h_t)}{\partial \omega_{h_t,a}} \frac{\partial \ln \pol(a_t \mid h_t)}{\partial \omega_{h_t,a'}}$ and that $\Pr_\mdp^\pol(h_t) =\sum_h \Pr_\mdp^\pol(h_t|h) \Pr_\mdp^\pol(h)= \sum_{h} \mathbb{I}_{h_t \in h}1/T \Pr_\mdp^\pol(h).$ Equation ~\eqref{eq:g1_grad} follows from the absolute value of equation~\eqref{eq:pol_grad} being bounded by one.

\begin{align}
	||G_2||_F & = ||\sum_h \util(h) \Pr_\mdp^\pol(h) \sum_t \nabla^2_{\pol} \ln \pol(a_t \mid h_t)||_F \\
	\le &T  ||\sum_h \Pr_\mdp^\pol(h) \sum_t \nabla^2_{\pol} \ln \pol(a_t \mid h_t)||_F \\
	\le &T  ||\sum_{h_t} T\Pr_\mdp^\pol(h_t) \nabla^2_{\pol} \ln \pol(a_t \mid h_t)||_F \\
	\le &T  \sqrt{\sum_{h_t} T^2\Pr_\mdp^\pol(h_t)^2 1} \\
	\le &T  \sqrt{T^2\sum_{h_t} \Pr_\mdp^\pol(h_t) 1} \\
	\le &T^2 
\end{align}
Similarly to the case for $G_1$, the steps follow the definition of the Frobenius norm, the observation that each element is weighted by $\Pr_\mdp^\pol(h_t) T$, and that the absolute value of the partial derivatives is bounded by 1.

Finally this yields 
\begin{equation}
l \le ||\nabla^2_{\pol} \util(\pol, \mdp)||_F \le  ||G_1||_F + ||G_2||_F \le  T^2( |\mathcal{A}| +1) . 
\end{equation}

$\eregret(., \beta)$ is $\mathcal{L}$-Lipschitz if $||\nabla_{\pol} \util(\pol, \mdp)||_2 \le \mathcal{L}$.
\begin{align}
||\nabla_{\pol} \util(\pol, \mdp)||_2
&=
||\sum_h \util(h) \Pr_\mdp^\pol(h) \sum_t \nabla_\pol \ln \pol(a_t \mid h_t) ||_2 \\
&\le  ||\sum_h \util(h) \Pr_\mdp^\pol(h) \sum_t \nabla_\pol \ln \pol(a_t \mid h_t) ||_F \\
&\le T   \sqrt{T^2 \sum_{h_t}   \Pr_\mdp^\pol(h_t)^2 1^2} \\
&\le  \max_h (|\util(h)|)  T.
\end{align}
This then gives $\mathcal{L}\le T^2$.

\end{proof}
\begin{proof}[Proof of Lemma~\ref{lemma:approximate_convex}.]
	Firstly, 
\begin{align}
\min_{\pi \in \Pols}\eregret(\pi, \bel^{\epsilon,*}) \\
\ge \min_{\pi \in \PolsEps}\eregret(\pi, \bel^{\epsilon,*}) - \epsilon \\
\ge \min_{\pi \in \PolsEps}\eregret(\pi, \bel^{*}) - \epsilon \\
\ge \min_{\pi \in \Pols}\eregret(\pi, \bel^{*}) - \epsilon
\end{align}
which completes the first part of the proof.

Secondly from the definition of c-convexity, and the fact that 
$\nabla_\beta  \min_{\pi \in \Pols}\eregret(\pi, \bel^{*})^T(\bel-\bel^*)$ must be zero since the gradient must be zero in any direction that does not move out of $\Bels$, 
we have
\begin{align}
	\min_{\pi \in \Pols}\eregret(\pi,\bel) \le \min_{\pi \in \Pols}\eregret(\pi, \bel^{*}) - c ||\bel^{*}-\bel||_2^2.
\end{align}
Rearranging and setting $\bel=\bel^{\epsilon,*}$ finishes the proof.
\end{proof}

  \section{Additional results for finite MDPs}  \label{appendix:finitemdp}
  \begin{figure}
  	\centering
  	\includegraphics[width=0.7\textwidth]{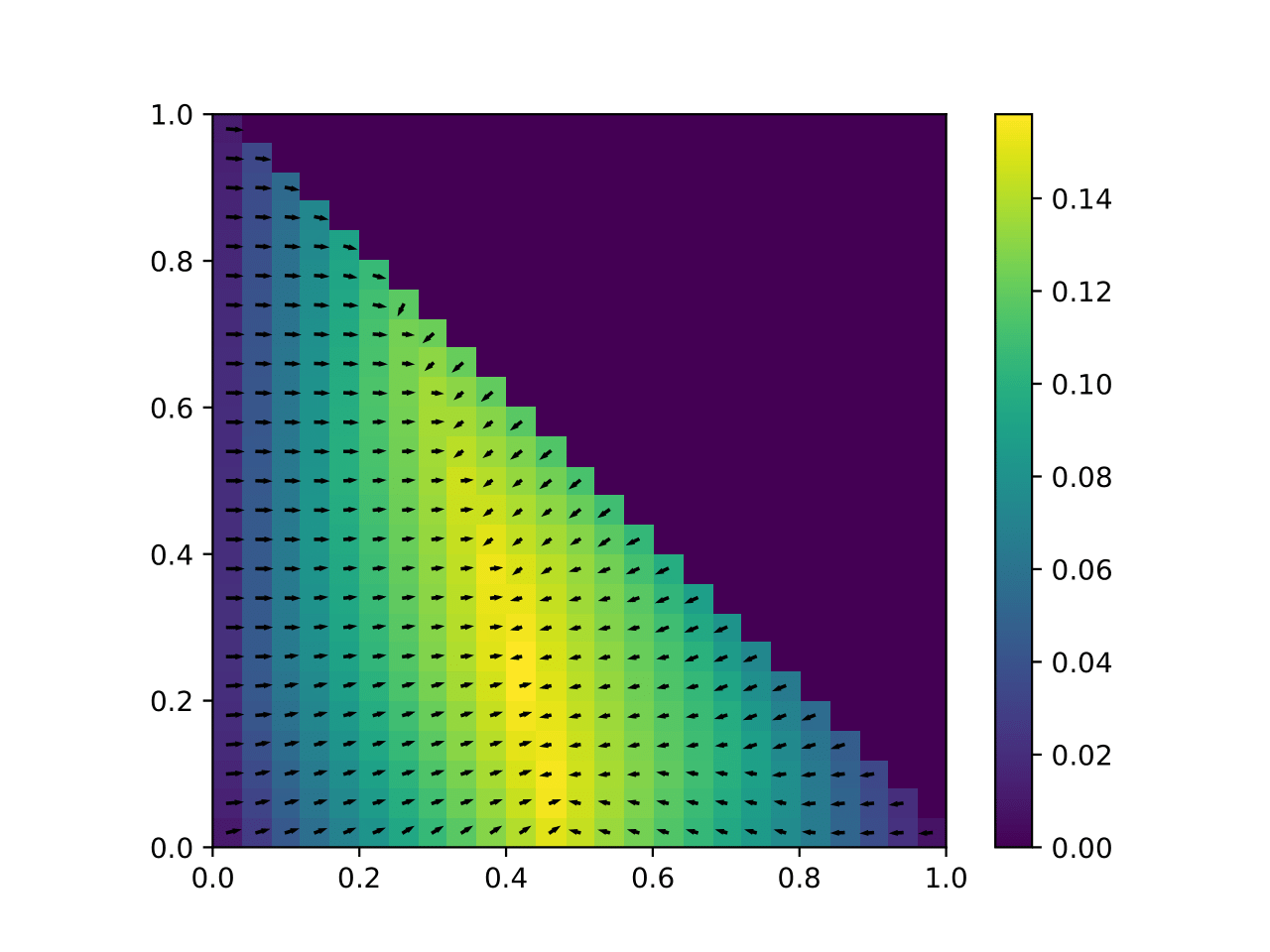}
  	\caption{Visualisation of Bayesian regret for three finite-horizon MDPs. The arrows show the gradients of the Bayesian regret for the corresponding Bayes-optimal policy. The axes represent the belief of two of the MDPs while the belief of the final MDP is given by 1-x-y.}
  	\label{fig:mdpgrid}
  \end{figure}
 
  In this section we generate MDPs as in the same way as in Section \ref{sec:finitemdp}, with the difference that Table \ref{tab:15mdp} uses $\gamma=0.9$.
  
  Figure \ref{fig:mdpgrid} gives an example of what the  Bayesian regret landscape looks like for a task with three MDPs. The change in  Bayesian regret for the fixed optimal policy of a certain belief is visualised with arrows.
  
  In Table \ref{tab:15mdp} we have some additional results comparing the performance of
  the uniform-prior and worst-case prior policies. In particular, we
  generate 5 sets of 16 MDPs. For each set, we calculate the minimax
  policy and the best response to the uniform prior. We then calculate
  the worst-case Bayesian regret for each policy. As we can expect,
  the minimax policy significantly outperforms the uniform best response policy.
\begin{table}[]
	\center
	\caption{Comparison of worst-case Bayesian regret for optimal policies at minimax and uniform belief for 16 MDP tasks.}
	\begin{tabular}{llllll}
		\toprule
		%seed& 123 & 234 & 345 & 456 & 567
		Seed & 1 & 2 & 3 & 4 & 5 \\
		\cmidrule{2-6}
		Minimax& 0.247 & 0.314 & 0.348  & 0.342 & 0.363 \\
		Uniform & 0.640  & 0.554 & 0.484  & 0.646 & 0.850 \\
		\bottomrule
	\end{tabular}
	\label{tab:15mdp}
\end{table}

\end{document}